\documentclass[sigconf]{acmart}

\usepackage{amsmath}
\usepackage{caption}
\usepackage{subcaption}
\usepackage{makecell}
\usepackage{multirow}
\usepackage[linesnumbered,ruled,vlined]{algorithm2e}

\renewcommand\footnotetextcopyrightpermission[1]{} 
\AtBeginDocument{%
  \providecommand\BibTeX{{%
    \normalfont B\kern-0.5em{\scshape i\kern-0.25em b}\kern-0.8em\TeX}}}

\setcopyright{acmcopyright}
\copyrightyear{2018}
\acmYear{2018}
\acmDOI{10.1145/1122445.1122456}

\acmConference[]{}{}{}
\acmBooktitle{}
\acmPrice{15.00}
\acmISBN{978-1-4503-XXXX-X/18/06}

\begin{document}

\title{Privacy-Aware Human Mobility Prediction via Adversarial Networks}

\author{Yuting Zhan, Hamed Haddadi}
\affiliation{Imperial College London
  \country{UK}
}
\author{Alex Kyllo, Afra Mashhadi}
\affiliation{University of Washington
  \country{USA}
}
\renewcommand{\shortauthors}{Yuting, Hamed and Afra, et al.}

\begin{abstract}

As various mobile devices and location-based services are increasingly developed in different smart city scenarios and applications, many unexpected privacy leakages have arisen due to geolocated data collection and sharing. 
While these geolocated data could provide a rich understanding of human mobility patterns and address various societal research questions, privacy concerns for users' sensitive information have limited their utilization.
In this paper, we design and implement a novel LSTM-based adversarial mechanism with representation learning to attain a privacy-preserving feature representation of the original geolocated data (\textit{i.e.}, mobility data) for a sharing purpose.
We quantify the utility-privacy trade-off of mobility datasets in terms of trajectory reconstruction risk, user re-identification risk, and mobility predictability.
Our proposed architecture reports a Pareto Frontier analysis that enables the user to assess this trade-off as a function of Lagrangian loss weight parameters. 
The extensive comparison results on four representative mobility datasets demonstrate the superiority of our proposed architecture and the efficiency of the proposed privacy-preserving features extractor.
Our results show that by exploring Pareto optimal setting, we can simultaneously increase both privacy (45\%) and utility (32\%).

\end{abstract}



\maketitle

\section{Introduction}

Geolocation and mobility data collected by location-based services (LBS)~\cite{huang2018location}, can reveal human mobility patterns and address various societal research questions~\cite{kolodziej2017local}. 
For example, Call Data Records (CDR) have been successfully used to provide real-time traffic anomaly and event detection~\cite{toch2019analyzing, wang2020deep}, and a variety of mobility datasets have been used in shaping policies for urban communities~\cite{ferreira_deep_2020} and epidemic management in the public health domain~\cite{oliver2015mobile,oliver2020mobile}.
Human mobility prediction based on users' trajectories, a popular and emerging topic, supports a series of important applications ranging from individual-level recommendation systems to large-scale smart transportation.

While there is no doubt of the usefulness of predictive applications for mobility data, privacy concerns regarding the collection and sharing of individuals’ mobility traces have prevented the data from being utilized to their full potential~\cite{shokri2011quantifying, beresford2003location, krumm2009survey}.  
A mobility privacy study conducted by De Montjoye et al~\cite{de2013unique} illustrates that four spatio-temporal points are enough to identify 95\% of the individuals in a certain granularity. 
As human mobility traces are highly unique, a mechanism capable of decreasing the user re-identification risk can offer enhanced privacy protection in mobility data sharing.

In the past decade, the research community has extensively studied privacy of geolocated data via various location privacy protection mechanisms (LPPM)~\cite{gedik2005location, gedik2007protecting}. 
Some traditional privacy-preserving approaches such as k-anonymity and geo-masking have shown to be insufficient to prevent users from being re-identified~\cite{de2013unique,song2010limits,gonzalez2008understanding, malekzadeh2020privacy}.
More recently, some related works also try to apply machine-learning or deep-learning based approaches to explore the effective LPPM. 
Rao et al proposed an LSTM-TrajGAN model to generate privacy-preserving synthetic mobility datasets for data sharing and publication~\cite{rao2020lstm}. 
Feng et al investigated human mobility data with privacy constraints via federated learning, achieving promising prediction performance while preserving the personal data on the local devices~\cite{feng2020pmf}. 
Though these state-of-the-art models provide a reasonable balance between utility and privacy, the effectiveness of the privacy mechanism and utility metrics have not been fully investigated. 

An appropriate and effective framework to allow researchers and practitioners to easily assess the trade-off between utility and privacy of mobility datasets at various granularity currently does not exist and can be highly impactful for the research community.

\begin{figure*}[t]
     \centering
     \begin{subfigure}[b]{0.72\textwidth}
         \centering
         \includegraphics[width=\textwidth]{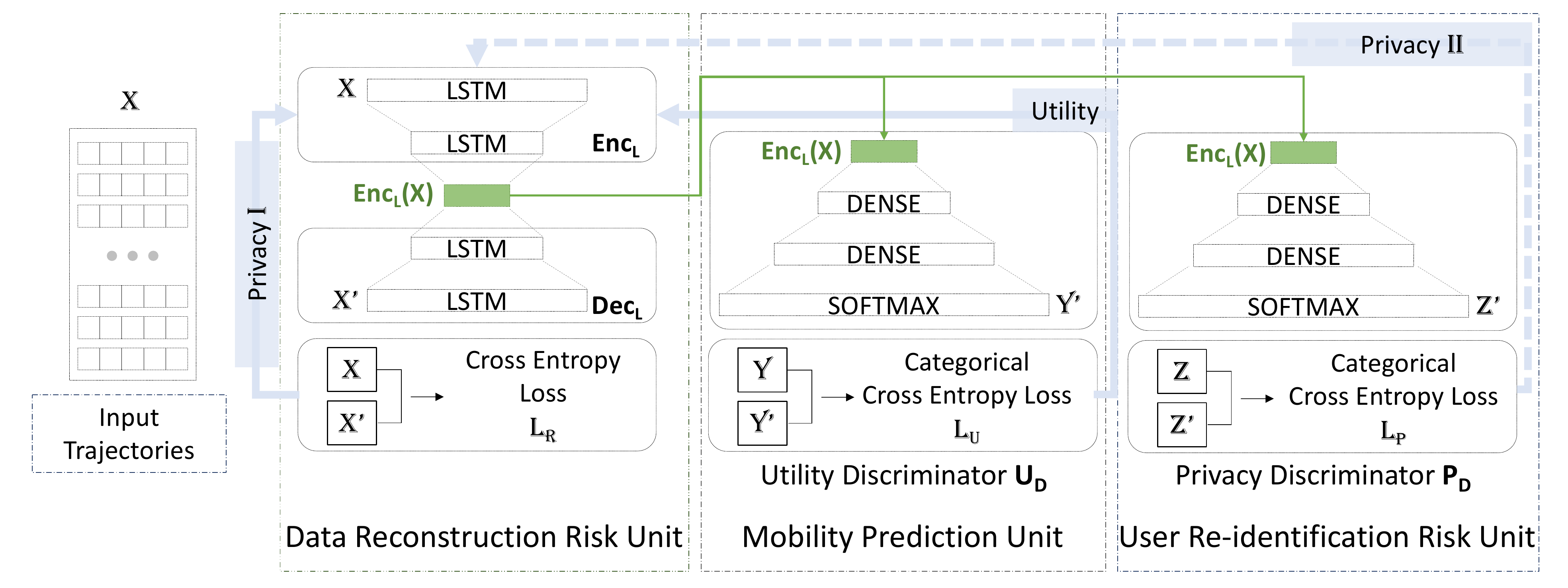}
         \caption{}
         \label{fig:proposed}
     \end{subfigure}
     \begin{subfigure}[b]{0.22\textwidth}
         \centering
         \includegraphics[width=\textwidth]{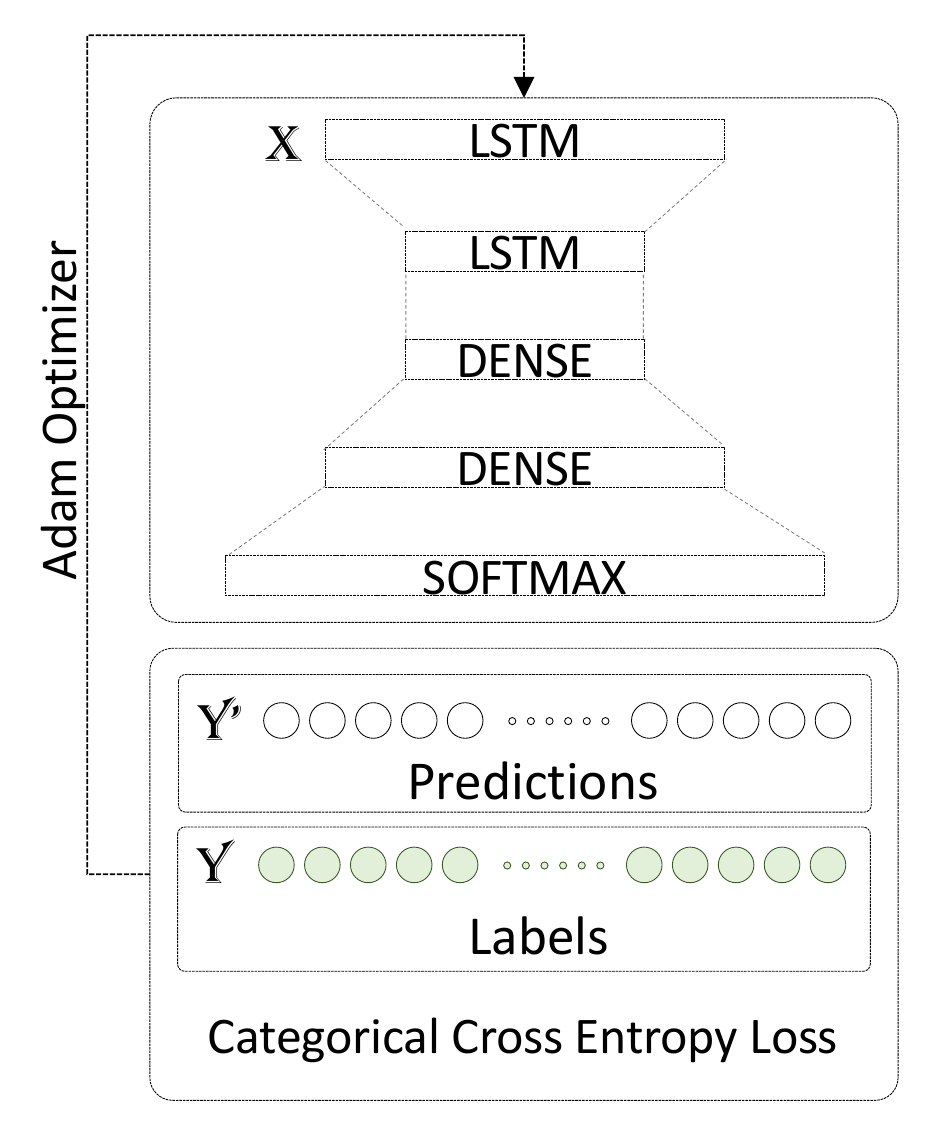}
         \caption{}
         \label{fig:baseline}
    \end{subfigure}
    \caption{(a) Schematic overview of the proposed privacy-preserving adversarial architecture with representation learning, consisting of data reconstruction risk unit, mobility prediction unit, and user re-identification risk unit. (b) The baseline LSTM network for standalone classifiers.}
    \Description{proposed architecture}
    \label{fig:architecture}
\end{figure*}

To this end, we posit an architecture for quantifying the utility-privacy trade-off of mobility datasets in terms of data reconstruction leakage (\textit{i.e.}, \textit{Privacy I}), user re-identification risk (\textit{i.e.}, \textit{Privacy II}), and mobility predictability (\textit{i.e.}, \textit{Utility}). 
In order to do so, we explore a novel mechanism to investigate these trade-offs and train a privacy-preserving feature extractor $Enc_L$ based on representation learning and adversarial learning. 
Inspired by PAN~\cite{liu2019privacy} (privacy adversarial network), we employ adversarial learning to better balance the potential trade-off between privacy and utility. 
In contrast to PAN, which focuses on the privacy of images, our approach is designed for complex time-series data that exhibits spatial-temporal characteristics. 
At the core of our architecture lies an LSTM auto-encoder (AE) with three branches, corresponding to the three training optimization objectives of the feature extractor $Enc_L$: i) to $maximize$ the loss associated with the reconstructed output by generative learning, ii) to $minimize$ the prediction loss using the learned representation from the $Enc_L$ by discriminative learning, and iii) to $maximize$ the percentage of users who are re-identifiable through their trajectories by discriminative learning. 
We use Lagrange multipliers to vary the weights that are given to each of these objectives before combining them into a total loss, $L_{sum}$.
The output of this model is a Pareto-Frontier analysis that would guide the user in investigating the trade-off between utility and privacy. 
 
We report the analysis of our architecture by a thorough evaluation on four real-world representative mobility datasets.
A benchmark comparison is carried out with the state-of-the-art algorithm based on Generative Adversarial Network (GAN)~\cite{goodfellow2014generative}, namely LSTM-TrajGAN ~\cite{rao2020lstm}. 
The results show that the synthetic dataset generated by TrajGAN is not {\em pareto-optimal} in all four cases. That is, in the given spatial-temporal granularity, it is possible to achieve a better privacy level for a dataset with the same utility value and vice versa.
The contributions of our work are the following:

\begin{itemize}

    \item We posit an LSTM-based location \textbf{p}rivacy protection mechanism via representation learning and \textbf{a}dversarial learning to learn a privacy-preserving feature extraction \textbf{e}ncoder, namely LSTM-PAE.
    
    \item We provide extensive analysis on different mobility tasks and quantify the privacy bound and utility bound of the target mobility dataset, along with a trade-off analysis between these contrasting objectives. 
    
    \item We compare our model over four real-world mobility datasets with a state-of-the-art GAN-based network that attempts to generate synthetic privacy-preserving mobility data.
    
    \item We offer the architecture as an open-source system to the researchers and practitioners.

\end{itemize}

The rest of this paper is structured as follows: Section \ref{DoF} describes our proposed LSTM-PAE model. In Section \ref{ExpS}, we describe the experimental settings for our reported results and provide a brief overview of the datasets used to evaluate our framework. Section \ref{MAME} reports an in-depth analysis of our proposed LSTM-PAE model. In Section \ref{FraE}, we demonstrate an evaluation of our framework over four mobility datasets and report on the utility-privacy trade-off of our proposed method in comparison with the state-of-the-art GAN based model. Finally, we review the related work in Section \ref{RelW} and conclude the paper with future work directions in Section \ref{Conl}.

\section{Design of the Architecture}
\label{DoF}

\subsection{Problem Definitions}
Before describing our proposed LSTM-PAE model in detail, we first give a brief problem definition of the trade-off between mobility data utility and privacy in terms of mobility prediction accuracy, user re-identification efficiency and data reconstruction differences.

\textbf{Data Utility:}
Mobility datasets are of great value for understanding human behavior patterns, smart transportation, urban planning, public health issue, pandemic management, and etc.
Many of these applications rely on the next location forecasting of individuals, which in the broader context can provide an accurate portrayal of citizens' mobility over time.
Mobility prediction not only can be analyzed to understand personalized mobility patterns, but can also inform the allocation of public resources and community services.
We focus on the capability of \emph{mobility prediction} (\emph{next location forecasting}) in this paper, and leverage the accuracy of the prediction as an important metric for quantifying the data utility.
Hence, the definition of \textit{Utility} is concluded as followed:

\textit{Utility} (\textit{U}): the mobility predictability (\textit{i.e.}, prediction accuracy)

\textbf{Privacy Protection:}
With more and more intelligent devices and sensors are utilized to collect information about human activities, the trajectories also expose increasing intimate details about users' lives, from their social life to their preferences. 
The capability of re-identification is important to balance the risks and benefits of mobility data usage, for all data owners, third parties, and researchers. 
We leverage the data reconstruction risk and user re-identification risk as our privacy metrics to evaluate our proposed privacy-aware architecture.
Hence, the definition of \textit{Privacy} is summarized as followed:

\textit{Privacy I} (\textit{PI}): the differences between the reconstructed data $X'$ and the original data $X$, that is, information loss in the reconstruction process.

\textit{Privacy II} (\textit{PII}): the user re-identification inaccuracy, that is, the user de-identification effectiveness.

\subsection{Architecture Overview}
\label{section: module overview}

Figure~\ref{fig:architecture} presents the basic workflow of the proposed privacy-preserving adversarial architecture with representation learning.
We aim to train an encoder $Enc_L$ to produce feature representations $f$ from the target mobility data by jointly optimizing these extracted feature weights using the combined losses of the data reconstruction risk unit, mobility prediction unit, and user re-identification risk unit simultaneously, during adversarial training.
The encoder $Enc_L$ is trained to achieve a better trade-off between predictability accuracy and user privacy budgets by extracting more information about the mobility predictability but less about the user privacy.
The \textit{utility} is measured via the mobility prediction task (\textit{i.e.}, prediction accuracy), and the \textit{privacy leakage} is quantified with two risks, listed as the data reconstructor risk (\textit{i.e.}, information loss) and the user re-identification risk (\textit{i.e.}, re-identification inaccuracy).
Hence, the proposed LSTM-PAE model is composed of three crucial units.
We discuss the functionality of each unit before we discuss the overall performance of the model.

\subsubsection{Architecture Composition Units}
\hfill

\textit{I. Mobility Prediction Unit}:  

\textit{i. Definition of Mobility Trajectory}: the raw geolocated data or other mobility data commonly contain three elements: user identifier \textit{u}, timestamps \textit{t}, location identifiers \textit{l}.
Hence, each location records \textit{r} could be denoted as \emph{$r_i$} = [\textit{$u_i$}, \textit{$t_i$}, \textit{$l_i$}], while each location sequence \emph{S} is a set of ordered location records \emph{$S_n$} =~\{\textit{$r_1$, $r_2$, $r_3$, $\cdots$, $r_n$}\}, namely \textit{mobility trajectory}.
Therefore, given the past mobility trajectory \emph{$S_n$} =~\{\textit{$r_1$, $r_2$, $r_3$, $\cdots$, $r_n$}\}, the mobility prediction task is to infer the most likely location \emph{l$_{n+1}$} at the next timestamp \emph{t$_{n+1}$}.
The data fed into the encoder $Enc_L$ are a list of trajectories with specific sequence length \emph{SL}, that is \{$S_{sl}^1$, $S_{sl}^2$, $S_{sl}^3$, $\cdots$, $S_{sl}^j$\}. 
For instance, if the sequence length is 10, that indicates each trajectory contains 10 history location records \textit{r}, \emph{$S_{10}$} = \{$r_1$, $r_2$, $r_3$, $\cdots$, $r_{10}$\}, and $SL=10$.

\textit{ii. Unit Design}: 
the mobility prediction unit is regarded as the \textit{utility discriminator} $U_D$ in the proposed architecture, as shown in Figure~\ref{fig:architecture}. 
This unit is composed of three parts, the input part with multi-modal embedding of trajectory information, the sequential part with LSTM layers, and an output part with the softmax activation function.
As per the aforementioned definition, the trajectories in this work are shown as location sequences \emph{S}. 
First, the location identifiers \textit{l} and timestamps \textit{t} are converted into one-hot vectors.
We then employ long short-term memory (LSTM)~\cite{hochreiter1997long} layers to model the mobility patterns and sequential transition relations in these mobility trajectories. 
As a prominent variant of the recurrent neural network, LSTM networks exhibit brilliant performance on modelling entire sequences of data, especially for learning long-term dependencies via gradient descent~\cite{zhan2019towards}.
Following the sequential module, the softmax layer outputs the probability distribution of the prediction results. This probability distribution is converted to the top-n accuracy metrics to illustrate the unit performance.

\textit{II. User Re-identification Risk Unit}: 

\textit{i. Definition of User Re-identification}: the privacy risk via user re-identification risk arises due to the high uniqueness of human traces~\cite{de2013unique}.
We assume each trajectory \emph{S} is originally labeled with a corresponding user identifier \textit{u}, and the user re-identification risk unit is to infer the user \textit{u} to whom the target trajectory \emph{$S_n$} = \{$r_1$, $r_2$, $r_3$, $\cdots$, $r_n$\} belongs. We thereby leverage the user identifiers as the ground-truth values for the user identity classes. 
However, this identity information is what we want to protect in the proposed adversarial network, that is, the extracted features $f$ should convey as little user identifiable information as possible, to decrease the user re-identification accuracy.

\textit{ii. Unit Design}:
the user re-identification risk unit is regarded as the \textit{privacy discriminator} $P_D$ in the proposed architecture. 
the unit is composed of three parts, the input part with one-hot embedding of user identity, the sequential part with LSTM layers, and an output part with softmax function.
First, the user identity list is converted into one-hot vectors. 
Similar to the mobility prediction unit, the user re-identification risk unit also applies LSTM layers to better extract the spatial and temporal characteristics of the context. 
A softmax function with cross-categorical entropy loss function is applied to 
output a categorical probability distribution of the user re-identification task.
We then use the top-n accuracy of this classifier as the metrics of user re-identification privacy risk (\textit{i.e.}, \textit{Privacy II}). 
The more accurately a classifier can re-identify the user when given a trajectory, the higher the risk of disclosing private data.

\textit{III. Data Reconstruction Risk Unit}:
this unit is the encoder $Enc_L$ unit in reverse. 
It is designed to evaluate the differences (\textit{i.e.}, \textit{Privacy I}) between the reconstructed data $X'$ and the original input data $X$.
A malicious party is free to explore any machine learning model and reconstruct the data if they have the shared extracted features $f$. 
We use a layer-to-layer reverse architecture of our encoder $Enc_L$ to build a strong \textit{Data Reconstruction Risk Unit}. 
We leverage \textit{Euclidean} and \textit{Manhattan} distance as our metrics to measure the differences between the $X$ and $X'$.

Following the basic discussion of three units, we present the overall design of the proposed LSTM-PAE model.

\subsubsection{Overall Design}
\hfill

\begin{table*}[t!]\small
    \centering
    \begin{tabular}{|c|cccc|cc|cccc|}
     \hline
\multirow{2}{*}{\shortstack{Dataset-City}} & \multicolumn{4}{c|}{Bounding Box} & \multicolumn{2}{c|}{Record Counts} & \multicolumn{4}{c|}{Number}\\
\cline{2-11}
         & \multicolumn{2}{c}{Latitude} & \multicolumn{2}{c|}{Longitude} & Train  & Test & User ID & POI & Weekday & Hour\\
         \hline
        MDC-Lausanne & 46.50 & 46.61 & 6.58 & 6.73 & 77393 & 19429 & 143 & 149 & 7 & 24\\
        Privamov-Lyon & 45.70 & 45.81 & 4.77 & 4.90 & 62077 & 16859 & 58 & 129 & 7 & 24\\
        GeoLife-Beijing & 39.74 & 40.07 & 116.23 & 116.56 & 95038 & 24578 & 145 & 960 & 7 & 24\\
        FourSquare-NYC & 40.55 &  40.99 & -74.28 & -73.68 & 43493 & 11017 & 466 & 1712 & 7 & 24\\
        \hline
    \end{tabular}
    \caption{Overview of four mobility datasets.}
    \label{tab:datasets}
\end{table*}

Our proposed \textbf{p}rivacy-preserving \textbf{a}dversarial feature \textbf{e}ncoder, the LSTM-PAE, is based on representation learning and adversarial learning and aims to ease data sharing privacy concerns. 
As shown in Figure~\ref{fig:proposed}, we design a multi-task adversarial network to learn an LSTM-based encoder $Enc_L$, 
which can generate the optimized feature representations $f=Enc_L(X)$ via lowering privacy disclosure risk of user identification information and improving the task accuracy (\textit{i.e.}, mobility predictability) concurrently.
Two potential malicious privacy leakages from the data reconstruction risk unit and the user re-identification risk unit, are attempted to retrieve sensitive information from the feature representations $f$.

Given mobility raw data $\mathcal{X}$ for \textit{Privacy I} (\textit{e.g.}, data reconstruction risk unit), the ground-truth label $z_i$ for \textit{Privacy II} (\textit{e.g.}, user re-identification risk unit), and the ground-truth label $y_i$ for utility (\textit{e.g.}, mobility prediction), we train the LSTM encoder $Enc_L$ of this multi-task network by adversarial learning to learn the representation $\mathcal{F}=Enc_L(\mathcal{X})$. 
We design a specific loss function, namely \textit{sum loss}  $\mathcal{L}_{sum}$, for this optimization process.
Specifically, when reconstructing the data $\mathcal{X'}$, an LSTM decoder $Dec_L$ attempts to recreate the data based on the features $\mathcal{F}$, that is $Dec_L: \mathcal{F}\rightarrow \mathcal{X'}$. This data reconstruction unit is trained by maximizing the reconstruction loss $\mathcal{L_R}$ while minimizing the $\mathcal{L}_{sum}$.
The mobility prediction unit, that is the utility discriminator $U_D$, is trained to output a probability distribution of the next location of interest, and this distribution has Y potential classes.
Discriminative training here is to maximize the prediction accuracy by minimizing the utility loss $\mathcal{L_U}$, denoted as $\min\mathcal{L_U}$. 
The user re-identification risk unit, that is the privacy discriminator $P_D$, is trained to re-identify whom the target trajectory belongs. 
Then in this privacy discriminator, the user re-identification loss $\mathcal{L_P}$ is maximized, denoted as $\max\mathcal{L_P}$.

In general, the encoder $Enc_L$ should satisfy high predictability (\textbf{\textit{min}}~$\mathcal{L_U}$) and low user re-identification accuracy (\textbf{\textit{max}}~$\mathcal{L_P}$) of the mobility data when maximizing the reconstruction loss (\textbf{\textit{max}}~$\mathcal{L_R}$) in reverse engineering, where the training objective can be written as (details is in Algorithm~\ref{algo_disjdecomp} attached in Appendix):
\begin{equation}
\begin{aligned}
    \text{min} \left(\mathcal{L}_{sum}\right) &= -\lambda_1 \left(\max\mathcal{L_R}\right) + \lambda_2 \left(\min\mathcal{L_U}\right) - \lambda_3 \left(\max\mathcal{L_P}\right) \\ 
    &= - \lambda_1  \| Dec_L(\mathcal{F}) - \mathcal{X}\|^2 + \lambda_2  (-\sum_{i=1}^{\mathbf{Y}} y_i\text{log}(U_D(\mathcal{F}))) \\
    &- \lambda_3  (-\sum_{i=1}^{\mathbf{Z}} z_i\text{log}(P_D(\mathcal{F})))
\end{aligned}
\label{equ:sumloss}
\end{equation}
where $y_i$ is the ground-truth label for \textit{Utility}, $z_i$ is the ground-truth value for \textit{Privacy II}; $\lambda_1$, $\lambda_2$ and $\lambda_3$ are Lagrange multipliers~\cite{beavis1990optimisation}.

The overall training is to achieve privacy-utility trade-off by adversarial learning on $\mathcal{L_R}$, $\mathcal{L_U}$, and $\mathcal{L_P}$ concurrently.
The gradient of the loss (\textit{i.e.}, $\theta_R$, $\theta_U$, $\theta_P$) back-propagates through the LSTM network to guide the training of the encoder $Enc_L$.
The encoder is updated with the \textit{sum loss} function $\mathcal{L}_{sum}$ until convergence.
The Lagrange multipliers are utilized to find the maxima or minima of a constrained problem. 
When two of them are set to zero, the model is transformed to a specific evaluation tool for a specific task.
When three multipliers are utilized together, they control the relative importance of each unit and guide the overall model to find the maxima or minima given the specific trade-off choices.

\section{Experimental Setting}
\label{ExpS}

\subsection{Datasets}
Experiments are conducted on four representative mobility datasets: Mobile Data Challenge Dataset (MDC)~\cite{laurila2012mobile}, PRIVA'MOV~\cite{mokhtar2017priva}, \\
FourSquare NYC~\cite{dingqi_yang_modeling_2015}, and GeoLife~\cite{zheng2011geolife}. Once imported into our architecture, each dataset was filtered and preprocessed individually to derive their respective test and training sets illustrated in Table~\ref{tab:datasets}. 

\noindent 
\textbf{MDC}: The MDC dataset, recorded from 2009 to 2011, contains a large amount of continuous mobility data for 184 volunteers with smartphones running a data collection software, in the Lausanne/Geneva area. 
Each record of the \textit{gps-wlan} dataset represents a phone call or an observation of a WLAN access point collected during the campaign ~\cite{laurila2012mobile}.  

\noindent 
\textbf{PRIVA'MOV}: The PRIVA'MOV crowd-sensing campaign took place in the city of Lyon, France from October 2014 to January 2016. Data collection was contributed by roughly 100 participants including university students, staff, and family members. The crowd-sensing application collected GPS, WiFi, GSM, battery, and accelerometer sensor data. For the purpose of this project, we only used the GPS traces from the dataset ~\cite{mokhtar2017priva}. 

\noindent
\textbf{GeoLife}: The GeoLife dataset was collected by Microsoft Research Asia from 182 users in the four and a half year period from April 2007 to October 2011 and contains 17,621 trajectories, mostly at a 5-second sampling rate \cite{zheng2011geolife}.

\noindent
\textbf{FourSquare NYC}: The Foursquare NYC dataset contains check-ins in NYC and Tokyo collected during the approximately 10 months from 12 April 2012 to 16 February 2013, containing 227,428 check-ins from 1,083 subjects in New York City \cite{dingqi_yang_modeling_2015}.

\begin{figure*}
     \centering
     \hfill
     \begin{subfigure}[b]{0.20\textwidth}
         \centering
         \includegraphics[width=\textwidth]{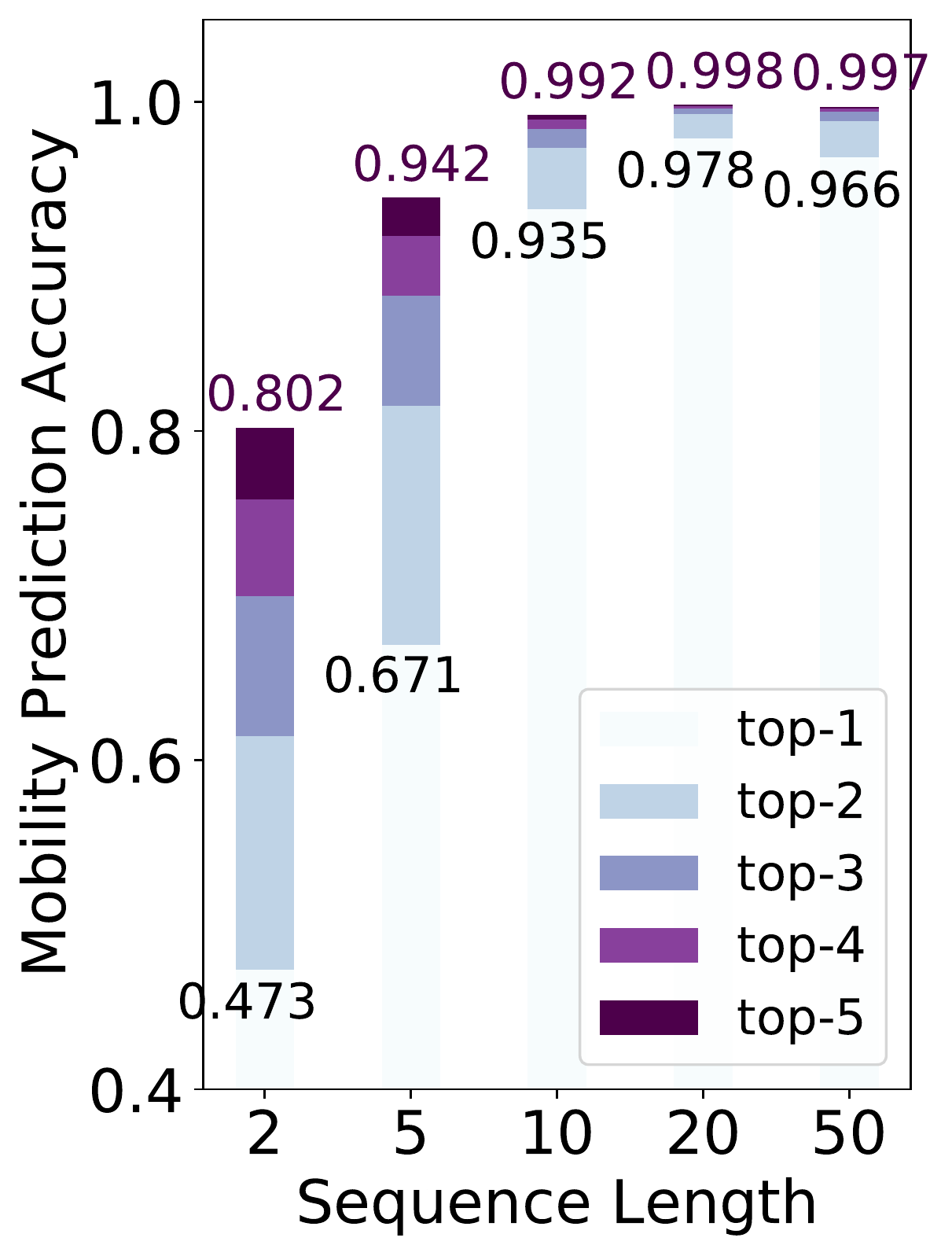}
         \caption{Utility discriminator $U_D$\\
         Performance on MDC}
         \label{fig:context_mdc_prediction}
     \end{subfigure}
     \hfill
     \begin{subfigure}[b]{0.20\textwidth}
         \centering
         \includegraphics[width=\textwidth]{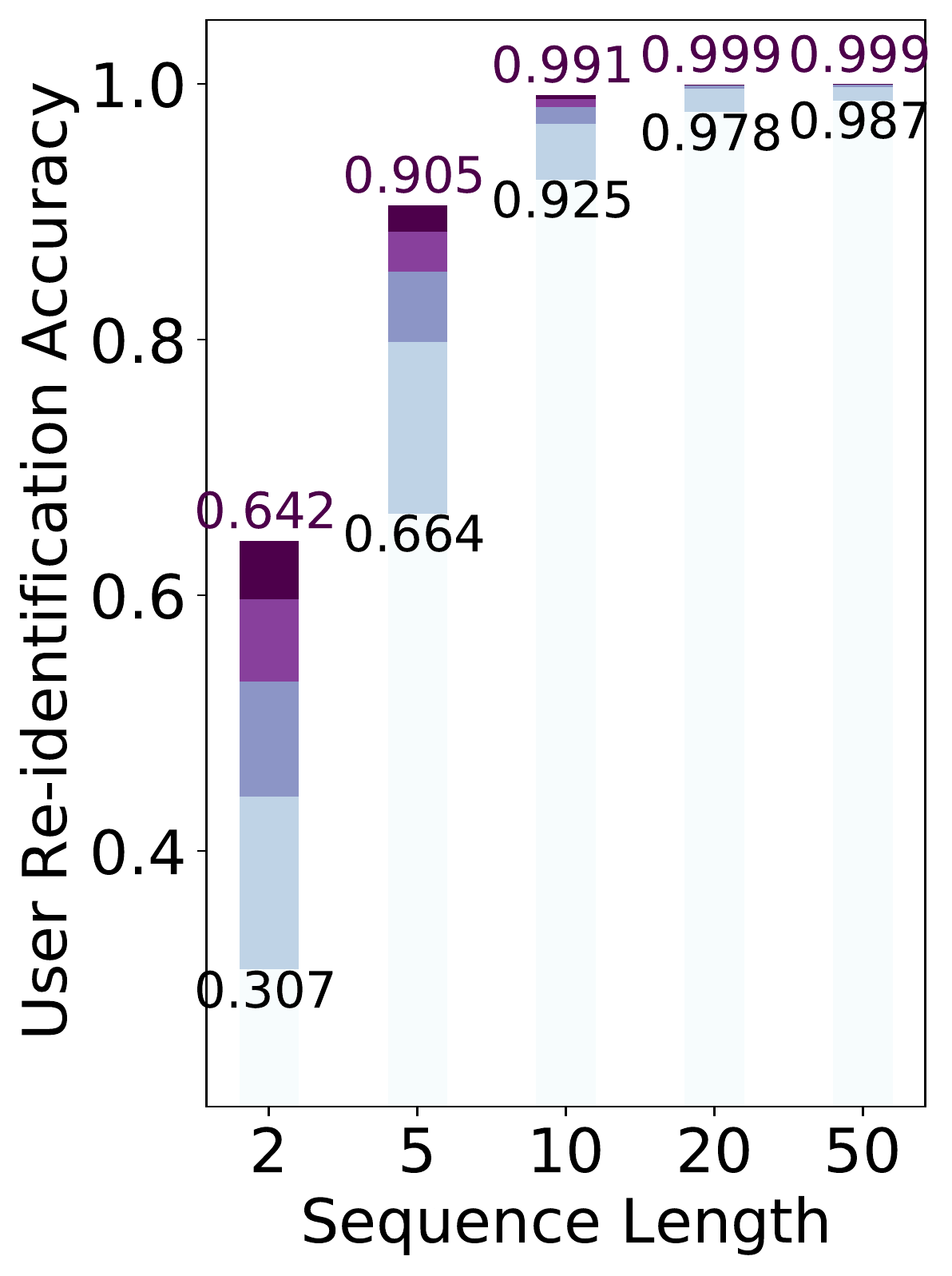}
         \caption{Privacy discriminator $P_D$
         Performance on MDC}
         \label{fig:context_mdc_recognition}
    \end{subfigure}
    \hfill
    \begin{subfigure}[b]{0.20\textwidth}
         \centering
         \includegraphics[width=\textwidth]{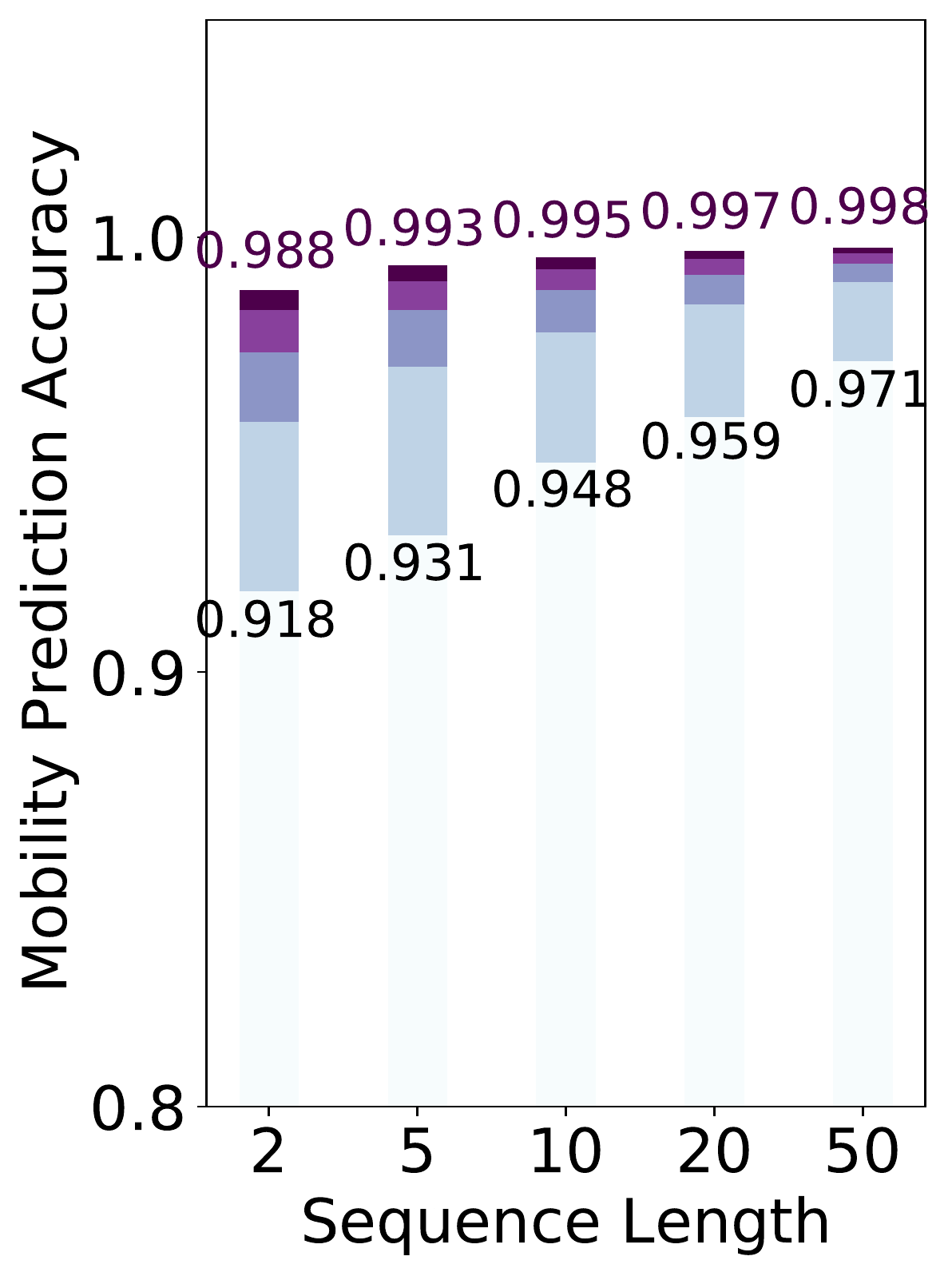}
         \caption{Utility discriminator $U_D$\\
         Performance on Priva'Mov}
         \label{fig:context_pri_prediction}
     \end{subfigure}
     \hfill
     \begin{subfigure}[b]{0.20\textwidth}
         \centering
         \includegraphics[width=\textwidth]{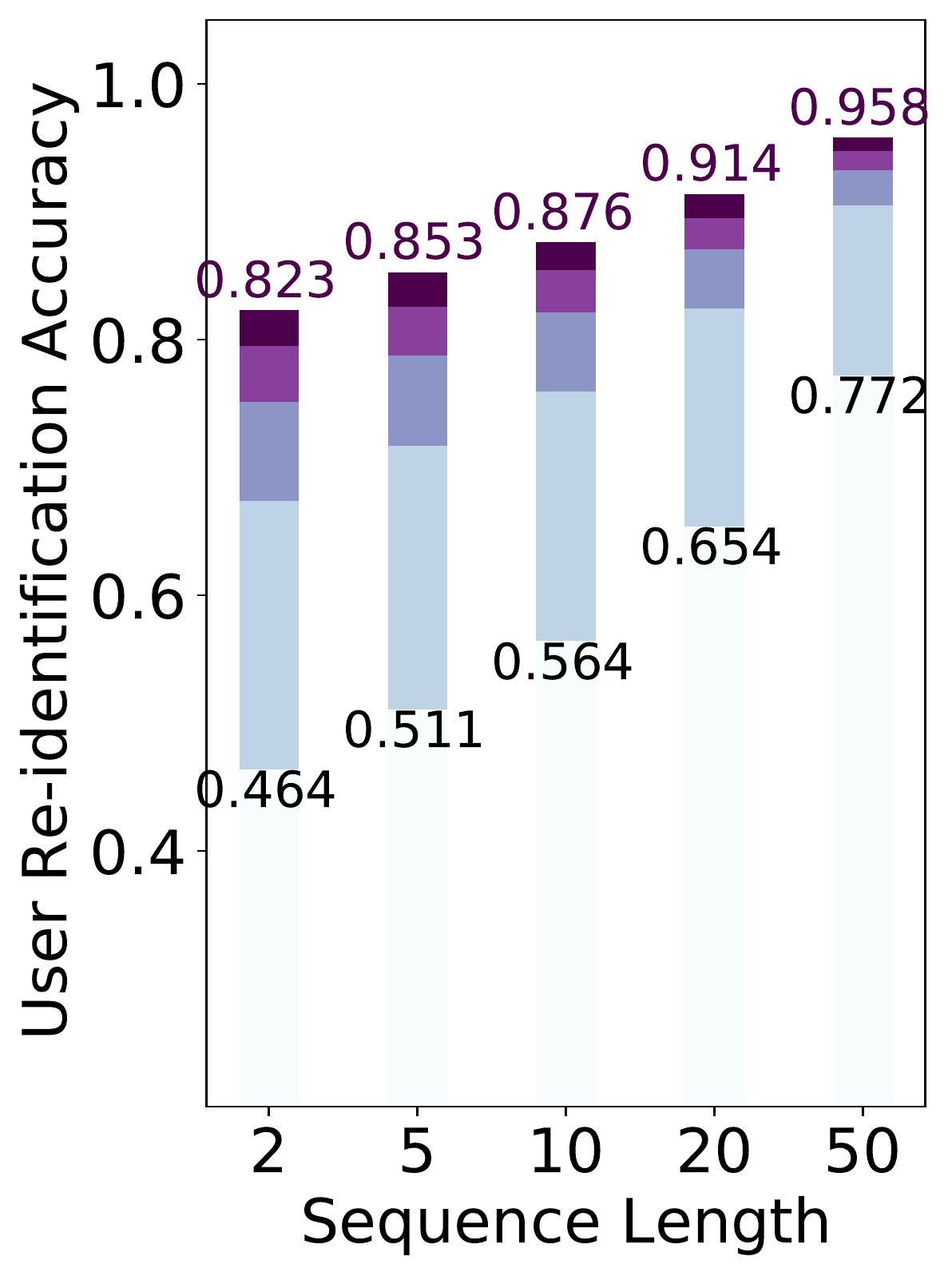}
         \caption{Privacy discriminator $P_D$
         Performance on Priva'Mov}
         \label{fig:context_mdc_recognition2}
    \end{subfigure}
    \hfill
    \caption{Mobility prediction accuracy and user re-identification accuracy change with the trajectory sequence length (SL) in our proposed mobility prediction unit and user re-identification unit. The color bars indicate the accuracy from top-1 to top-5, the black texts indicate the top-1 accuracy and the purple texts indicate the top-5 accuracy. For instance, the top-1 mobility prediction accuracy on MDC with \textit{SL} = 2 is 0.473, and the top-5 one is 0.802.}
    \Description{The effect of the context length}
    \label{fig:context}
\end{figure*}

\subsection{Training}
\subsubsection{Training of LSTM-PAE}
The main goal of the proposed adversarial network is to learn an efficient feature representation based on the utility and privacy budgets, using all users' mobility histories.
In most experiments in this work, the trajectory sequences consist of 10 historical locations with timestamps (\textit{i.e.}, $SL=10$), and the impact of the varying sequence lengths is discussed in Section~\ref{section:impact sl}.
After the pre-processing of the datasets, 70\% of the records of each user are segmented as the training dataset, and the remaining 30\% as the testing dataset. 
We utilize the mini-batch learning method to train the model until the expected convergence.
We take a gradient step to optimize the \textit{sum loss} $L_{sum}$ (\textit{i.e.}, Equation \ref{equ:sumloss}) in terms of $L_R$, $L_U$, and $L_P$ concurrently.
Meanwhile, the \textit{sum loss} $L_{sum}$ is also optimized by using the Adam optimizer.
All the experiments were performed with the Tesla V100 GPU; a round of training would take 30 seconds on average and each experiment trains for 1000 rounds.

\subsubsection{Training of the Comparison Model}
To provide a state of the art trained model for comparison, we re-implement the LSTM-TrajGAN model described in \cite{rao2020lstm}
using the same hyper-parameters, setting latent vector dimension to 100, using 100 LSTM units per layer, a batch size of 256, utilizing the Adam optimizer with learning rate 0.001 and momentum 0.5, and training for 200 epochs
(where one epoch is a pass through the entire training set).
We train LSTM-TrajGAN independently on the training split of each benchmark mobility dataset, and then use it to generate synthetic trajectories from the test set.
Then we train the proposed LSTM-PAE on the same training data and use it to generate a feature extraction from the same test data.
Finally, we evaluate the performance of the user re-identification unit and mobility prediction unit on the real and synthetic test sets generated by LSTM-TrajGAN, and compare the changes in accuracy to assess the relative utility and privacy of the TrajGAN and PAE.

\subsection{Metrics}
We set \textit{Euclidean}~\cite{ball1960short} and \textit{Manhattan} distance~\cite{black1998dictionary} as our evaluation metrics for the data reconstruction unit to evaluate the quality of the reconstructed data $X'$ generated from extracted features $f$.
\textit{Euclidean} distance gives the shortest or minimum distance between two points, while \textit{Manhattan} distance applies only if the points are arranged in the form of a grid, and both definitions are feasible for the problem we are working on. Note that these two distances have limited capability in showing the quality of the reconstructed data $X'$, however, they intuitively capture the differences between the original data $X$ and the reconstructed data $X'$. 

We leverage the top-n accuracy as our evaluation metric for both mobility prediction and user re-identification risk units. The accuracy of the mobility prediction unit is one of the most important factors in evaluating the utility of the extracted feature representation \textit{f}, where predictability of the \textit{f} increases as much as it can during the adversarial training.
On the other hand, the competing training objective is to decrease the accuracy of the user re-identification unit to enhance the privacy of \textit{f}.
The top-n metric computes the number of times where the correct label appears among the top $n$ labels predicted. The top-n metric takes n predictions with higher probability into consideration and it classifies the prediction as correct if one of them is a true label.
The top-1 to top-5 accuracies~\cite{zhan2019towards} are leveraged in our paper to discuss the performance of the proposed model.

\section{Experiments}
\label{MAME}

In this section, we first discuss the impact of the varying sequence length and varying Lagrange multipliers on the composition units before presenting the overall performance of the proposed architecture in next section. 

\subsection{Impact of Varying Sequence Lengths}
\label{section:impact sl}
The performance of utility discriminator $U_D$ (\textit{i.e.,} mobility prediction unit) and the privacy discriminator $P_D$ (\textit{i.e.,} user re-identification risk unit) exert great impact on the overall performance of the proposed LSTM-PAE. 
The trajectory length is the most important factor which could affect these units' performance.
We use two representative datasets (\textit{i.e.}, MDC and Priva'Mov) to present the impact of the varying sequence length on both discriminators. 

By changing the lengths of trajectory sequence $SL$ from 1 ($SL=1$) to 50 ($SL=50$), we observe that the length of trajectory sequence has a high impact on different tasks' accuracy (\textit{i.e.,} mobility prediction accuracy for $U_D$ and user re-identification accuracy for $P_D$) of two different datasets, while the impact in the MDC dataset is much higher than in the Priva'Mov dataset, as shown in the Figure~\ref{fig:context}. 
Comparing the Figure~\ref{fig:context_mdc_prediction} and Figure~\ref{fig:context_pri_prediction}, there is a much sharper increase on the MDC dataset. 
More specifically, when the sequence length is increased from 2 to 20, the top-1 mobility prediction accuracy on MDC increases from 0.473 to 0.978 (\textit{i.e.}, +50.5\%), while accuracy on Priva'Mov increases from 0.918 to 0.959 (\textit{i.e.}, only +4.1\%).
Similarly, more rapid growth appears in the user re-identification accuracy on MDC, which is +68.0\%, while the increase for Priva'Mov is only +30.8\%.
We conclude that the mobility predictability and user re-identification accuracy of a dataset might have a special link.
The mobility predictability of Priva'Mov is very high, almost higher than 90\%, but the user re-identification accuracy is always lower than 80\%, which also means the uniqueness of trajectories in this dataset is low. This low uniqueness suggests that the users in this dataset might share similar daily routes, which is reasonable, as we know these trajectories are collected from students at the same university.
For the MDC dataset, when $SL=10$, the user re-identification accuracy is quite high, indicating that the locations are more sparse in this dataset. However, the mobility predictability of this dataset is also high, which also emphasizes that this sparseness does not affect the predictability.
These phenomena indicate that the deep training of mobility predictability and user re-identification might share similar extracted features, while our proposed architecture attempts to extract features more suitable for mobility predictability but less suitable for user re-identification.

We note that the varying trajectory sequence length not only exerts impacts on the model performance, but also has a great influence on the computation time. 
For instance, the computation time at $SL=50$ costs six times as much as the one at $SL=5$. The computation time also varies between datasets. 
Hence, an appropriate choice of trajectory sequence length can avoid time-consuming computation and to achieve expected task inference accuracy. 
In our work, we place greater focus on the trajectory sequence lengths ranging from 5 to 10, which exhibits great performance in both the $U_D$ and $P_D$ while also keeps a low computation time.

\begin{figure}
     \centering
     \begin{subfigure}[b]{0.40\textwidth}
         \centering
         \includegraphics[width=\textwidth]{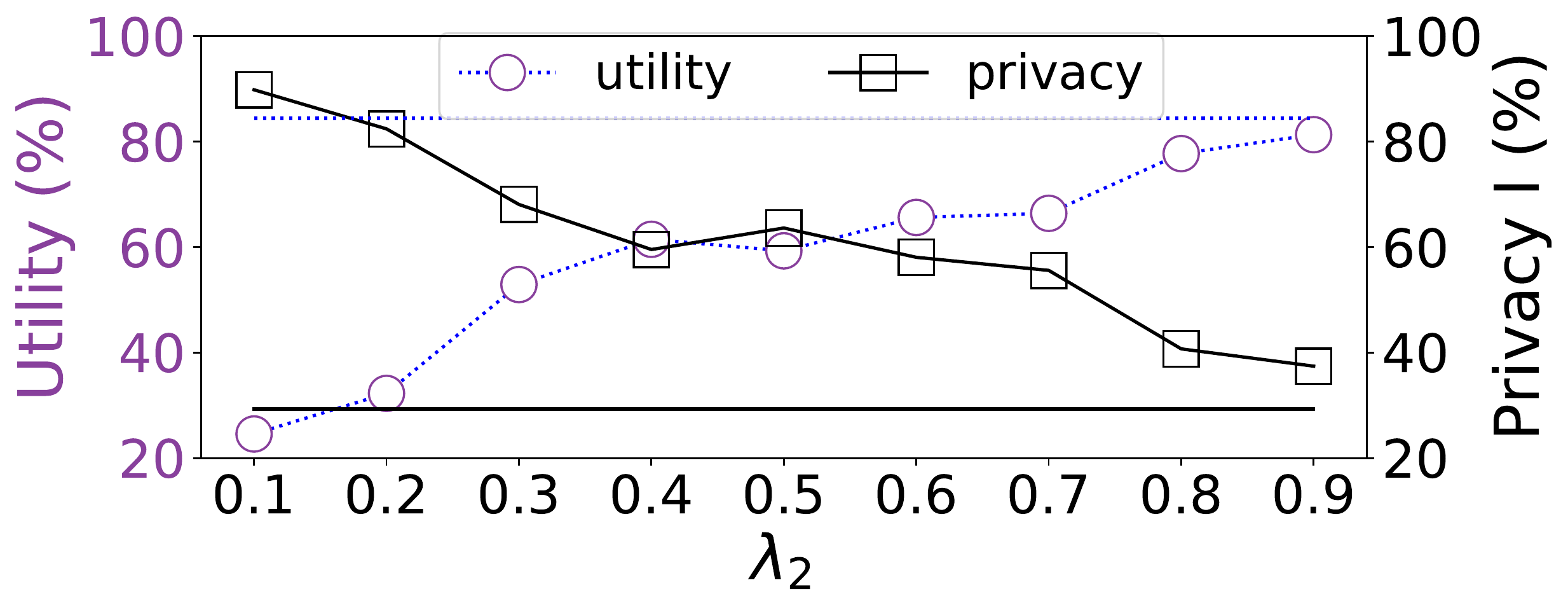}
         \caption{$\lambda_1 = 1-\lambda_2,~\lambda_3 = 0$}
         \label{fig:lambda3_zero_privacy}
     \end{subfigure}
     \begin{subfigure}[b]{0.40\textwidth}
         \centering
         \includegraphics[width=\textwidth]{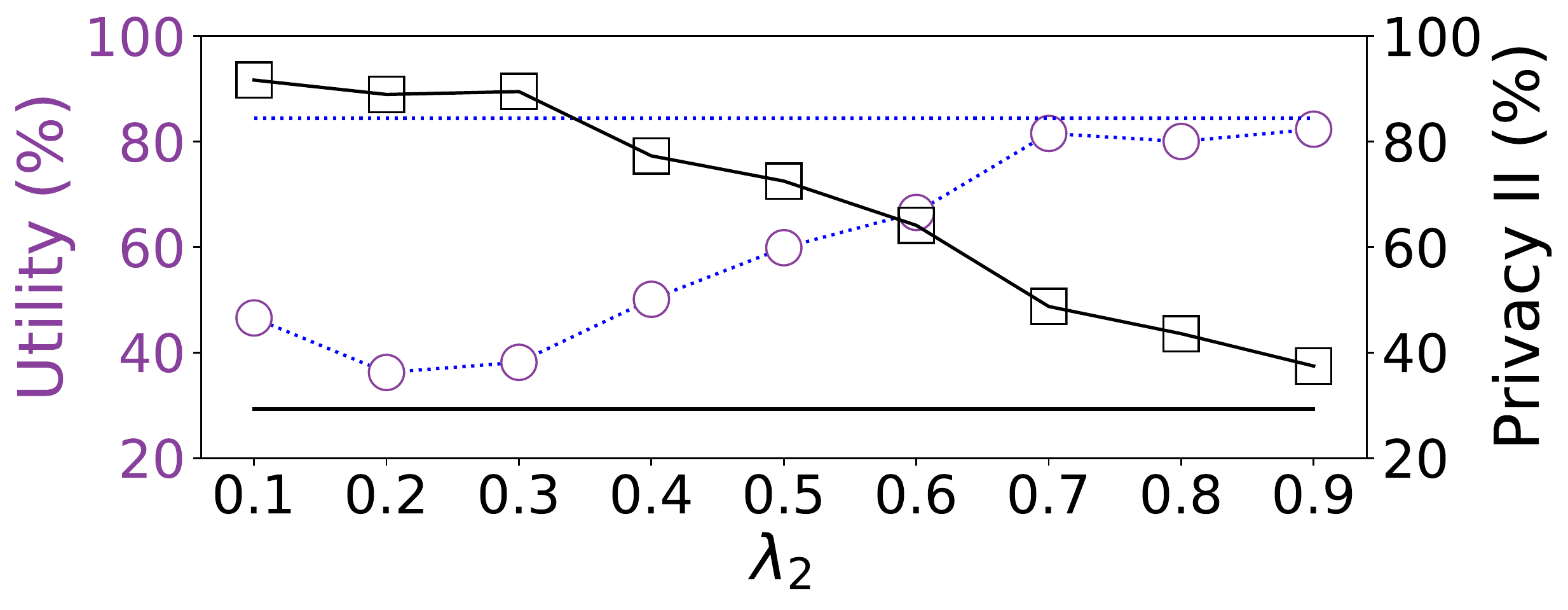}
         \caption{$\lambda_3 = 1-\lambda_2,~\lambda_1 = 0$}
         \label{fig:lambda1_zero_privacy}
    \end{subfigure}
    \caption{The utility-privacy trade-offs can be tuned by varying Lagrange multipliers. The primary y-axis (dashed line) represents utility, the secondary y-axis (solid line) represents privacy. The x-axis represents the value of the target Lagrange multiplier. }
    \Description{privacy and accuracy}
    \label{fig:lagranian}
\end{figure}

\subsection{Impact of Varying Lagrange Multipliers}
As we discussed in Section~\ref{section: module overview}, the \textit{sum loss function} $L_{sum}$ is a linear combination of $L_R$, $L_U$, and $L_P$ based on Lagrange multipliers. 
In other words, the loss optimization in our proposed architecture LSTM-PAE could be reframed as a Lagrangian optimization question. 
We evaluate the influence of different combinations of Lagrangian multipliers $\lambda_1$, $\lambda_2$, and $\lambda_3$ to the proposed adversarial network, as the results shown in Figure~\ref{fig:lagranian}.

In order to illustrate the adversarial effect between $L_R$ and $L_U$, as shown in the Figure~\ref{fig:lambda3_zero_privacy}, we compare the overall model performance in terms of the \textit{Utility} and \textit{Privacy I} by fixing the $\lambda_3 = 0$, and varying the other two multipliers by subjecting to $\lambda_1 = 1- \lambda_2$. Figure~\ref{fig:lambda1_zero_privacy} represent the effect between $L_P$ and $L_U$ by setting the $\lambda_1 = 0$, and compare the performance in terms of the \textit{Utility} and \textit{Privacy II}.
We could observe in both settings that the utility increases with a larger $\lambda_2$, which means when the mobility prediction unit is given more weight in the LSTM-PAE model, it would exert a positive impact on the data utility. 
We conclude that the utility-privacy trade-offs could be tuned by varying the Lagrange multipliers; the results in the Figure~\ref{fig:lagranian} also verify the effectiveness of our adversarial architecture.
Hence, in the next section we provide an architecture evaluation to aid in achieving a comprehensive and practical trade-off among \textit{Utility}, \textit{Privacy I} and \textit{Privacy II}, and use it to discuss the overall performance of the proposed LSTM-PAE.
We note that the balance of three units is far more complicated than the balance of two.
From the extensive experiment we conducted, initialing $\lambda_1=0.1$, $\lambda_2=0.6$, $\lambda_3=0.3$ can guide the model achieve the tradeoff most efficiently.

\section{Architecture Evaluation}
\label{FraE}

In this section, we present the comparison results of the proposed architecture LSTM-PAE and two baseline models under the same training setting. 

\textit{Baseline Models}

\textit{I. Standalone Model}: It comprises three independent sub-models, namely data reconstruction sub-model, mobility prediction sub-model, and user re-identification sub-model. 
Each of the sub-models have a similar layer design as the corresponding unit in the LSTM-PAE, however, the results of the three sub-models are completely independent and have no effect on others. 
Differently from the LSTM-PAE, which leverages adversarial learning to finally attain an extracted feature representation $f$ that satisfies the utility requirements and privacy budgets simultaneously, the standalone models only are trained for optimal inference accuracy at the individual tasks.

\textit{II. LSTM-TrajGAN}~\cite{rao2020lstm}: It is an end-to-end deep learning model to generate synthetic data which preserves essential spatial, temporal, and thematic characteristics of the real trajectory data. Compared with other common geomasking methods, TrajGAN can better prevent users from being re-identified.
While the mobility prediction ability of the synthetic data is not in their utility metrics, the TrajGAN work claims to preserve essential spatial and temporal characteristics of the original data, verified through statistical analysis of the generated synthetic data distributions, which is in a line with the mobility prediction based utility assessment in our work.
Hence, we train a mobility prediction model for each dataset and evaluate the mobility predictability of synthetic data generated by the TrajGAN.
In contrast to the TrajGAN that aims to generate synthetic data, our proposed LSTM-PAE is training an encoder $Enc_L$ that forces the extracted representations \textit{f} to convey maximal utility while minimizing private information about user identities, via adversarial learning.

\begin{table*}[t]\small
   \centering
   \begin{tabular}{ccccccccccc}
    \hline
    \multirow{2}{*}{\shortstack{Datasets}} & \multirow{2}{*}{\shortstack{Models}} & & \multicolumn{2}{|c|}{Privacy I} & \multicolumn{3}{c|}{Utility (\% for decline)} & \multicolumn{3}{c}{Privacy II (\% for gain)} \\
    \cline{4-11}
        &  & & \multicolumn{1}{|c}{Euc(log)} & \multicolumn{1}{c|}{Man(log)} & top-1  & top-3  & \multicolumn{1}{c|}{top-5} & top-1  & top-3 & top-5 \\
        \hline
        
        \multirow{3}{*}{\shortstack{MDC}} & Standalone & & 0.001 & 0.002 & 0.9347  & 0.9837  & 0.9922 & 0.9247 & 0.9819 & 0.9911\\
        & TrajGAN & & 3.526 & 5.456  & -46.32\% & -24.16\% & -15.98\% & +20.32\% & +8.13\% & +4.02\% \\
        & \multirow{2}{*}{\shortstack{Our Model}} & I & 2.294 & 4.542 & -54.56\% & -34.74\% & -25.10\% & \textbf{+69.80\%} & \textbf{+50.44\%} & \textbf{+39.95\%} \\
        & & II & \textbf{3.732} & \textbf{6.023} & \textbf{-13.43\%} & \textbf{-6.26\%} & \textbf{-3.95\%} & \textbf{+65.51\%} & \textbf{+45.11\%} & \textbf{+34.86\%} \\
        \hline
        
        \multirow{3}{*}{\shortstack{Priva'Mov}} & Standalone & & 1.281 & 2.554 & 0.9482  & 0.9878 & 0.9954 & 0.5643 & 0.8215 & 0.8765 \\
        & TrajGAN & & 3.704 & 5.779 & -6.60\% & -1.89\% & -0.93\% & +14.17\% & +14.35\% & +8.88\%\\
        & \multirow{2}{*}{\shortstack{Our Model}} & I & 1.740 & 3.433 & \textbf{-3.36\%} & \textbf{-1.59\%} & \textbf{-0.81\%} & +27.02\% & +14.19\% & +9.19\% \\
        & & II & \textbf{4.164} & \textbf{5.803} & -10.81\% & -6.83\% & -4.91\% & \textbf{+35.29\%} & \textbf{+14.97\%} & \textbf{+10.05\%} \\
        \hline
        
        \multirow{3}{*}{\shortstack{Geolife}} & Standalone & & 1.903 & 3.804 & 0.4705 & 0.6842 & 0.7636 & 0.6572 & 0.8690 & 0.9294\\
        & TrajGAN &  & 4.581 & 6.680 & -62.31\% & -50.45\% & -43.72\% & \textbf{+66.73\%} & \textbf{+47.89\%} & \textbf{+37.22\%} \\
        & \multirow{2}{*}{\shortstack{Our Model}} & I & 1.776 & 3.469 & -31.45\% & -25.02\% & -21.90\% & +54.88\% & +39.59\% & +30.81\% \\
        & & II & \textbf{4.616}& \textbf{6.928}& \textbf{-21.13\%} & \textbf{-18.78\%} & \textbf{-17.11\%} & +55.49\% & +40.40\% & +32.34\% \\
        
        \hline
        \multirow{3}{*}{\shortstack{FourSquare}} & Standalone & & 2.357 & 4.464 & 0.6468 & 0.8210 & 0.8823 & 0.8780 & 0.9735 & 0.9892 \\
        & TrajGAN & & \textbf{4.795} & \textbf{6.710} & -26.30\% & -22.30\% & -18.75\% & \textbf{+51.86\%} & +32.49\% & +23.49\% \\
        & \multirow{2}{*}{\shortstack{Our Model}} & I & 2.418 & 4.533 & -51.05\% & -41.45\% & -35.20\% & +53.47\% & +35.26\% & +25.86\% \\
        & & II & 4.541 & 6.638 & \textbf{-2.54\%} & \textbf{-3.14\%} & \textbf{-2.84\%} & \textbf{+51.08\%} & \textbf{+34.39\%} & \textbf{+26.16\%}\\
        \hline
        
    \end{tabular}
    \caption{Performance comparison between our proposed models with standalone model and the TrajGAN model. The \textit{Model I} is our proposed architecture without Lagrange multipliers, and the \textit{Model II} is the one with multipliers ($\lambda_1$ = 0.1, $\lambda_2$ = 0.8, and $\lambda_3$ = 0.1). The results shown in this table are all with trajectory sequence length 10 (\textit{i.e.}, \textit{SL} = 10). 
    The \textit{Privacy I} intuitively shows the difference between the raw data and reconstructed data; the \textit{Utility} (\%) represents the utility declines; and the \textit{Privacy II}(\%) represents the privacy gains calculated via the user re-identification \textit{inaccuracy} rate. 
    }
    \label{tab:compare}
\end{table*}

\subsection{Performance Comparison}
We first compare our proposed models with the standalone model and the LSTM-TrajGAN model on four representative mobility datasets, as details shown in Table~\ref{tab:compare}. 
The overall performance is evaluated in terms of the \textit{utility level} provided by the mobility prediction unit and the \textit{privacy threat} provided by two risk units 
The \textit{Model I} is our proposed architecture but without applying the Lagrange multipliers (\textit{i.e.}, where each losses are weighed equally), and the \textit{Model II} is the one with Lagrange multipliers (\textit{i.e.}, $\lambda_1=0.1,\ \lambda_2=0.8,\ \lambda_3=0.1$ for the results in Table~\ref{tab:compare}).
The results in Table~\ref{tab:compare} are based on the input data with trajectory sequence length 10 (that is $SL=10$). 
Because the standalone models are trained without the consideration for the utility-privacy trade-offs, the results on the standalone models can be leveraged to explain the best inference accuracy (\textit{i.e.}, mobility prediction accuracy and user re-identification accuracy) that each composition unit could achieve.
Differently from the standalone model, the TrajGAN and LSTM-PAE are both taking the utility-privacy trade-offs into consideration and we compare their trade-offs with the standalone version. Hence, in Table~\ref{tab:compare}, these results are shown in utility decline and privacy gain, both of which are in a percentage format.
The similarity indexes are leveraged to intuitively represent the difference between the original data $X$ and reconstructed data $X'$, where the larger value indicates numerical differences between them. 

Table~\ref{tab:compare} demonstrates that our proposed models, especially the one with Lagrangian multipliers, outperforms the LSTM-TrajGAN model across various datasets.
For instance, when models are trained with the MDC dataset, our \textit{Model II} achieves the best privacy-utility trade-offs among different models, as the utility decline is only 13.43\% but with 65.51\% privacy gain, while 46.32\% utility decline and 20.32\% privacy gain with the TrajGAN. 
The similarity indexes also indicate the reconstructed data $X'$ via \textit{Model II} has bigger differences than the one via the TrajGAN. 
Second, although the utility decline of the TrajGAN on the Priva'Mov dataset is 4.21\% higher than our \textit{Model II}, both two privacy metrics of the TrajGAN are worse than the \textit{Model II}. Our model has better overall trade-offs in utility requirements and privacy budgets. 
The performance on Geolife and FourSquare are similar but inverse, where the utility of our model is better than TrajGAN and with slightly weaker privacy preservation. 
We leverage the composite metrics to score the overall performance of different models, and demonstrates that our model achieves better utility-privacy trade-offs.
The comparisons between \textit{Model I} and \textit{Model II} also illustrate the importance of the Lagrange multipliers in not only providing flexibility to our proposed architecture that enable its application in different scenarios, but also enhancing the utility-privacy trade-offs in this special case.

\begin{figure*}
     \centering
     \begin{subfigure}[t]{0.8\textwidth}
     \centering
         \includegraphics[width=\textwidth]{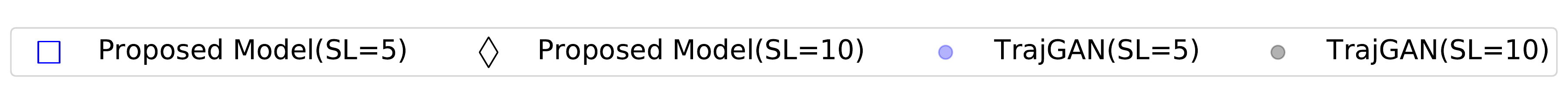}
     \end{subfigure}
     
     \begin{subfigure}[t]{0.2\textwidth}
         \centering
         \includegraphics[width=\textwidth]{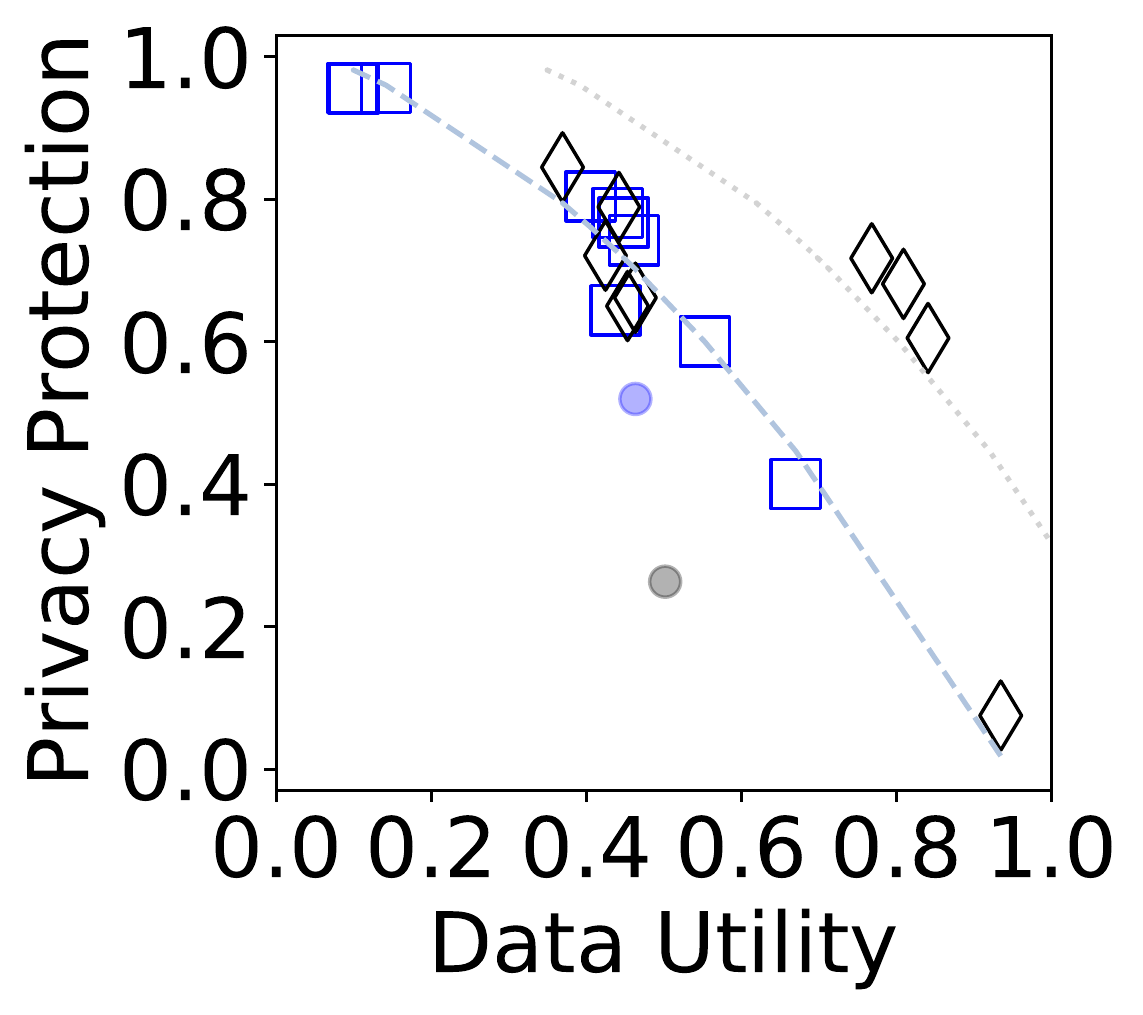}
         \caption{MDC}
         \label{fig:mdc}
     \end{subfigure}
     \begin{subfigure}[t]{0.2\textwidth}
         \centering
         \includegraphics[width=\textwidth]{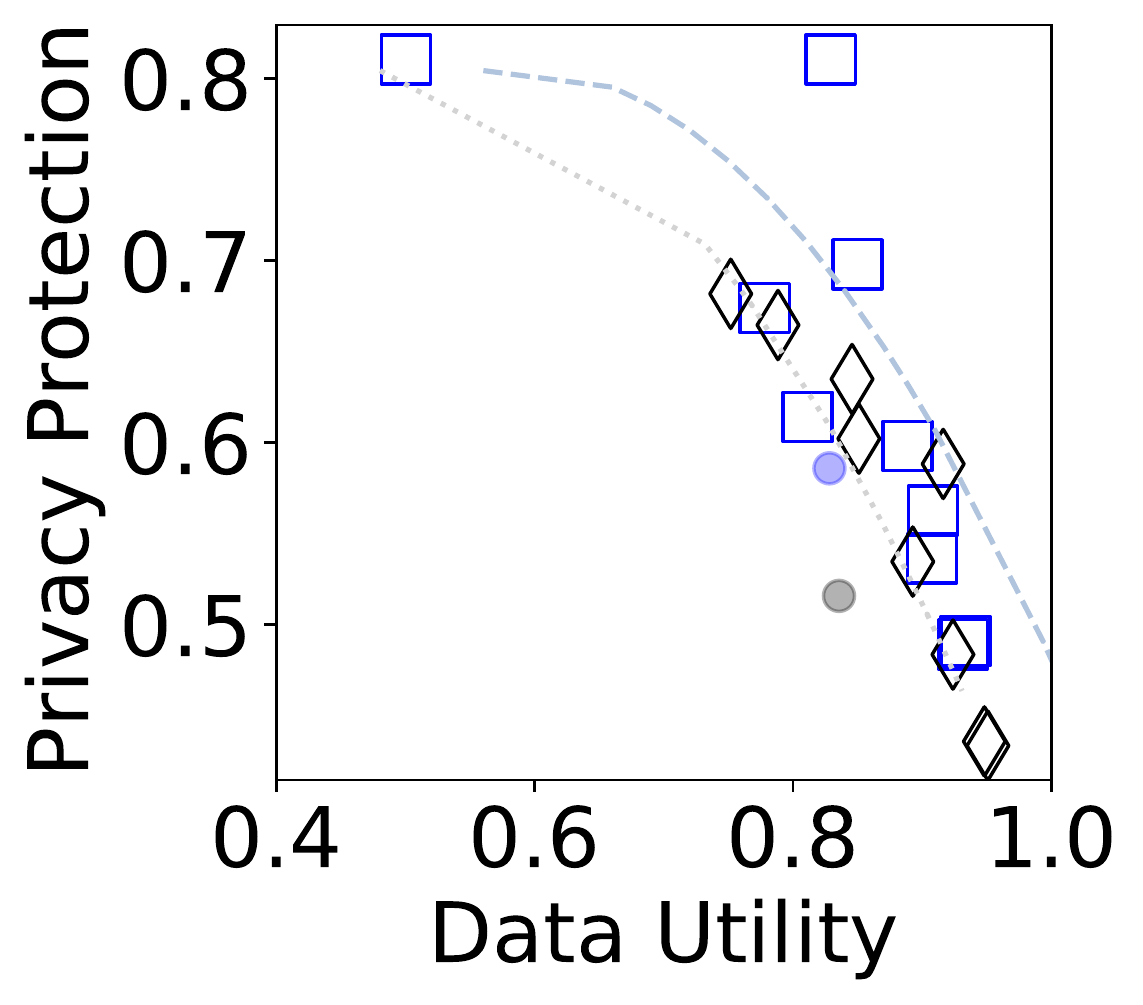}
         \caption{Priva\'Mov}
         \label{fig:privamov}
    \end{subfigure}
    \begin{subfigure}[t]{0.19\textwidth}
         \centering
         \includegraphics[width=\textwidth]{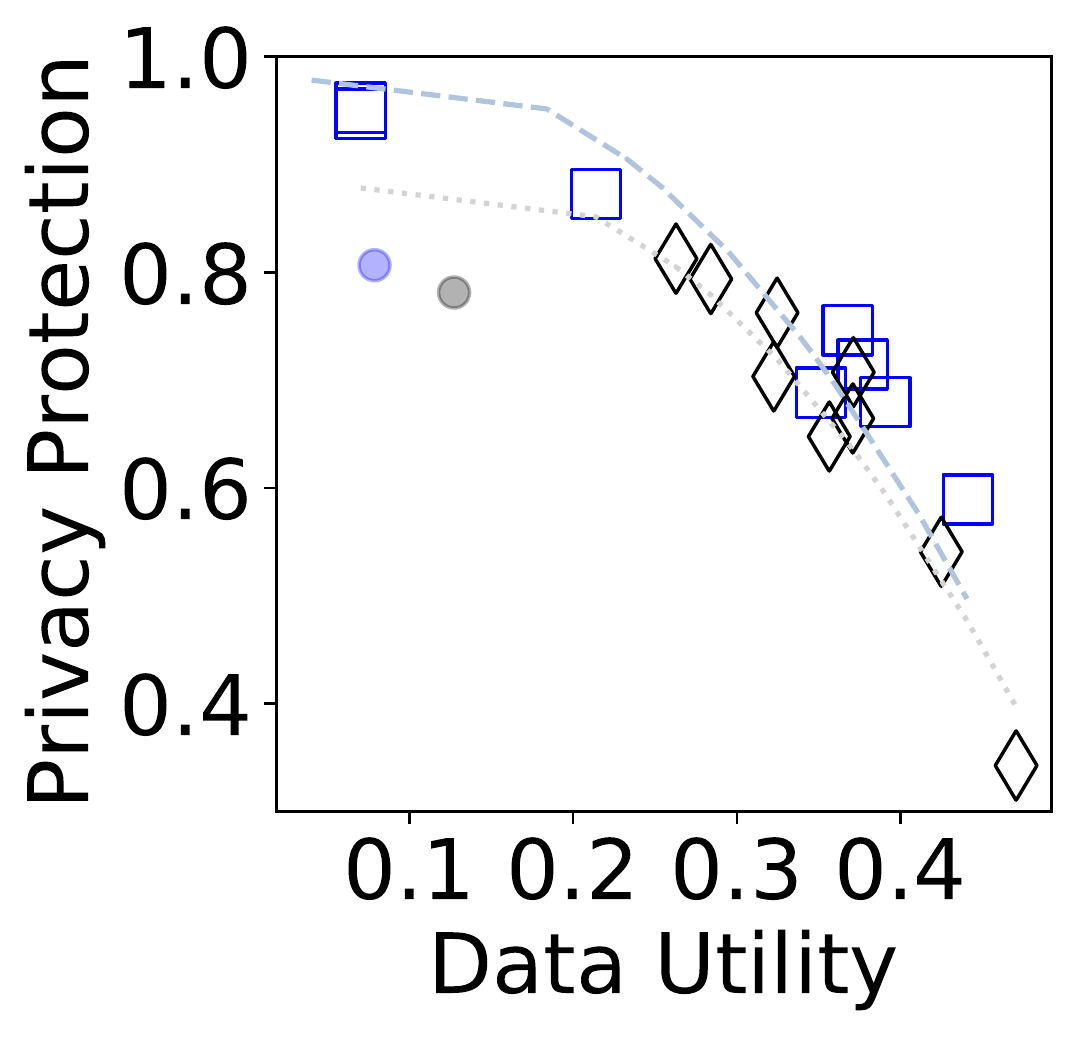}
         \caption{Geolife}
         \label{fig:geolife}
     \end{subfigure}
     \begin{subfigure}[t]{0.19\textwidth}
         \centering
         \includegraphics[width=\textwidth]{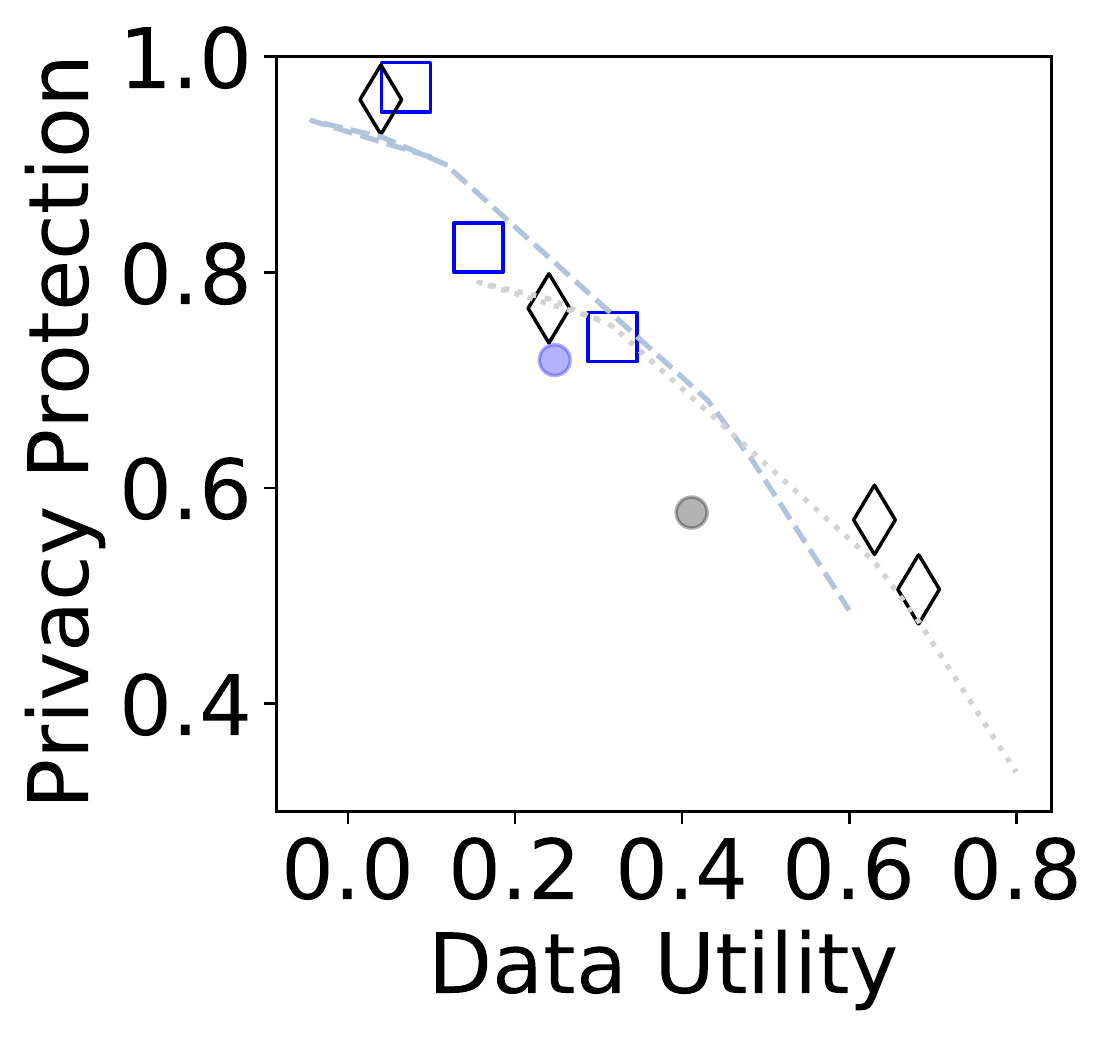}
         \caption{Foursquare}
         \label{fig:foursquare}
    \end{subfigure}
    \caption{Pareto Frontier trade-off of Utility and Privacy on four datasets. The hollow squares and diamonds present results of the proposed models.
    The solid points present results of the TrajGAN.
    Blue color means \textit{SL} = 5. Black color means \textit{SL} = 10.}
    \Description{}
    \label{fig:compare}
\end{figure*}

\begin{table*}[t]\small
   \centering
   \begin{tabular}{ccccccccccccc}
    \hline
    \multirow{2}{*}{\shortstack{Settings}} & \multicolumn{3}{|c|}{MDC} & \multicolumn{3}{c|}{Priva\'Mov} & \multicolumn{3}{c|}{Geolife} & \multicolumn{3}{c}{FourSquare} \\
    \cline{2-13}
        &  \multicolumn{1}{|c}{Euc} & Man & \multicolumn{1}{c|}{Utility}  & Euc & Man & \multicolumn{1}{c|}{Utility}  & Euc & Man & \multicolumn{1}{c|}{Utility}  & Euc & Man & Utility \\
        \hline
        I &  \textbf{+3.814} & \textbf{+6.122} & \textbf{-0.565} & +1.800 & +1.230 & -0.160 & \textbf{+1.451} & \textbf{+0.845} & \textbf{-0.208} & \textbf{+1.045} & \textbf{+0.580} & \textbf{-0.608} \\
        II & +3.804 & +6.120 & -0.493 & \textbf{+1.839} & \textbf{+1.125} & \textbf{-0.196} & +1.444 & +0.837 & -0.186 & +1.032 & +0.569 & -0.574 \\
        III & +3.746 & +6.049 & -0.166 & +1.999 & +1.277 & -0.097 & +1.444 & +0.834 & -0.146 & +0.988 & +0.535 & -0.406 \\
        IV & +3.732 & +6.023 & -0.125 & +2.251 & +1.272 & -0.102 & +1.426 & +0.821 & -0.099 & +0.926 & +0.487 & -0.016\\
        V & \textbf{+3.722} & \textbf{+6.012} & \textbf{-0.094} & \textbf{+1.829} & \textbf{+1.186} & \textbf{-0.055} & \textbf{+1.407} & \textbf{+0.809} & \textbf{-0.098} & \textbf{+0.902} & \textbf{+0.472} & \textbf{+0.036} \\
        \hline
        
    \end{tabular}
    \caption{Data reconstruction privacy leakage risk (\textit{PI}) versus utility decline (\textit{U}) on four mobility datasets. 
    We list five different settings of the Lagrange multipliers to discuss the potential range of the utility-privacy tradeoffs.
    Both "Euc" and "Man" are calculated by the log base 10 of the Euclidean and Manhattan distances.}
    \label{tab:euc}
\end{table*}

\begin{figure*}[t]
     \centering
     \includegraphics[width=0.85\textwidth]{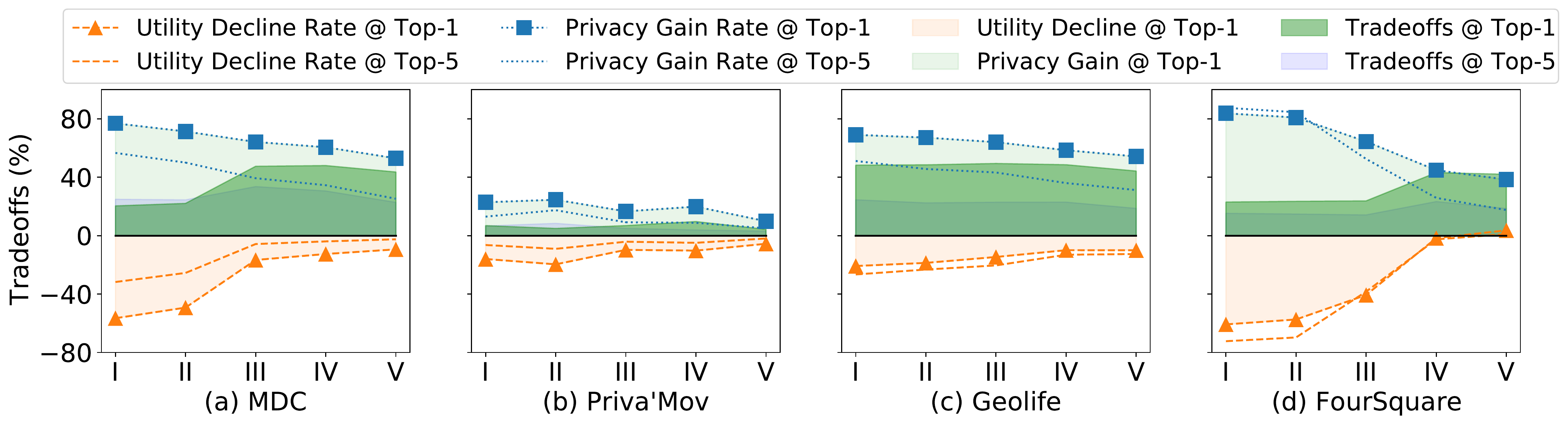}
     \caption{User re-identification privacy gain (\textit{PII}) versus utility decline (\textit{U}) on four datasets. 
     The orange area represents the utility decline while the light-green area represents privacy gain. The dark-green area represents the trade-offs between utility achievement and privacy budgets. The x-axis shows five different settings of the model, and the y-axis shows the trade-offs.}
    \Description{}
    \label{fig:utility-privacy}
\end{figure*}

\subsection{Utility-Privacy Analysis}
\label{UPA}

In this section, we first present the utility-privacy trade-off analysis between TrajGAN and our proposed LSTM-PAE in terms of the mobility prediction accuracy (\textit{i.e.}, \textit{U}) and user de-identification efficiency (\textit{i.e.}, \textit{PII}). 
We then discuss two privacy risks (\textit{i.e.}, \textit{PI} and \textit{PII}) of our proposed framework among four representative mobility datasets.

\subsubsection{Trade-off Comparison}
Figure~\ref{fig:compare} presents the trade-off comparisons of the four datasets in terms of the \textit{U} and \textit{PII} under different Lagrangian settings, where the \textit{hollow squares} and \textit{hollow diamonds} show the tradeoffs provided by the proposed LSTM-PAE in $SL=5$ and $SL=10$, respectively. 
The \textit{solid points} present the utility-privacy trade-off provided by the TrajGAN under the same spatial granularity and same trajectory sequence length. 
As can be seen from these results, in all four cases the synthetic dataset generated by the TrajGAN is not {\em pareto-optimal}. That is, in that given spatial-temporal granularity, the proposed architecture is able to achieve a better privacy level for a dataset with the same utility value. 
Compared with the TrajGAN, our proposed architecture improves utility and privacy at the same time on four datasets. Especially for the performance on the MDC dataset, the privacy improves 45.21\% than the TrajGAN while the utility also increases 32.89\%. 
These results illustrate that our proposed model achieves promising performance in training a privacy-sensitive encoder $Enc_L$ for different datasets.

\subsubsection{Privacy Leakage Risk Analysis}
After evaluating the superior performance of our proposed framework, we discuss the privacy leakage risks among four representative mobility datasets in terms of data reconstruction loss (\textit{PI}, \textit{"Euc"} and \textit{"Man"} in Table~\ref{tab:euc}) and user re-identification inaccuracy (\textit{PII}, \textit{privacy gain} in Figure~\ref{fig:utility-privacy}). We use five different combinations of Lagrangian multipliers to evaluate the comprehensive performance of the proposed model, namely setting \textit{I, II, III, IV, and V} in the Table~\ref{tab:euc} and Figure~\ref{fig:utility-privacy}.

Table~\ref{tab:euc} shows the impact of the data reconstruction privacy leakage risk (\textit{PI}) on the data utility decline (\textit{U}).
The \textit{"Euc"} and \textit{"Man"} in this table are calculated by the \textit{Euclidean} and \textit{Manhattan} distances, shown as the logarithm to the base 10 in the Table~\ref{tab:euc}, which intuitively demonstrates the difference between the reconstructed data $X'$ and the original one $X$. 
The results show that the impact of reconstructed data loss on the utility is high, thereby emphasizing the importance of balancing the trade-off between them.
Indeed, more information loss implies less privacy leakage risk, which also results in a larger utility decline.

\begin{figure*}[t]
     \centering
     \begin{subfigure}[b]{0.98\textwidth}
     \centering
         \includegraphics[width=\textwidth]{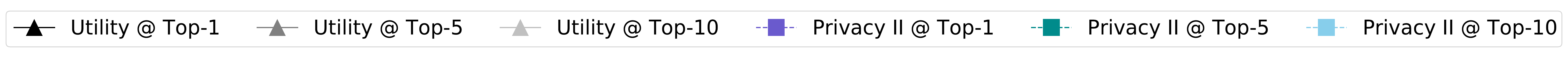}
     \end{subfigure}
     
     \hfill
     \begin{subfigure}[b]{0.24\textwidth}
         \centering
         \includegraphics[width=\textwidth]{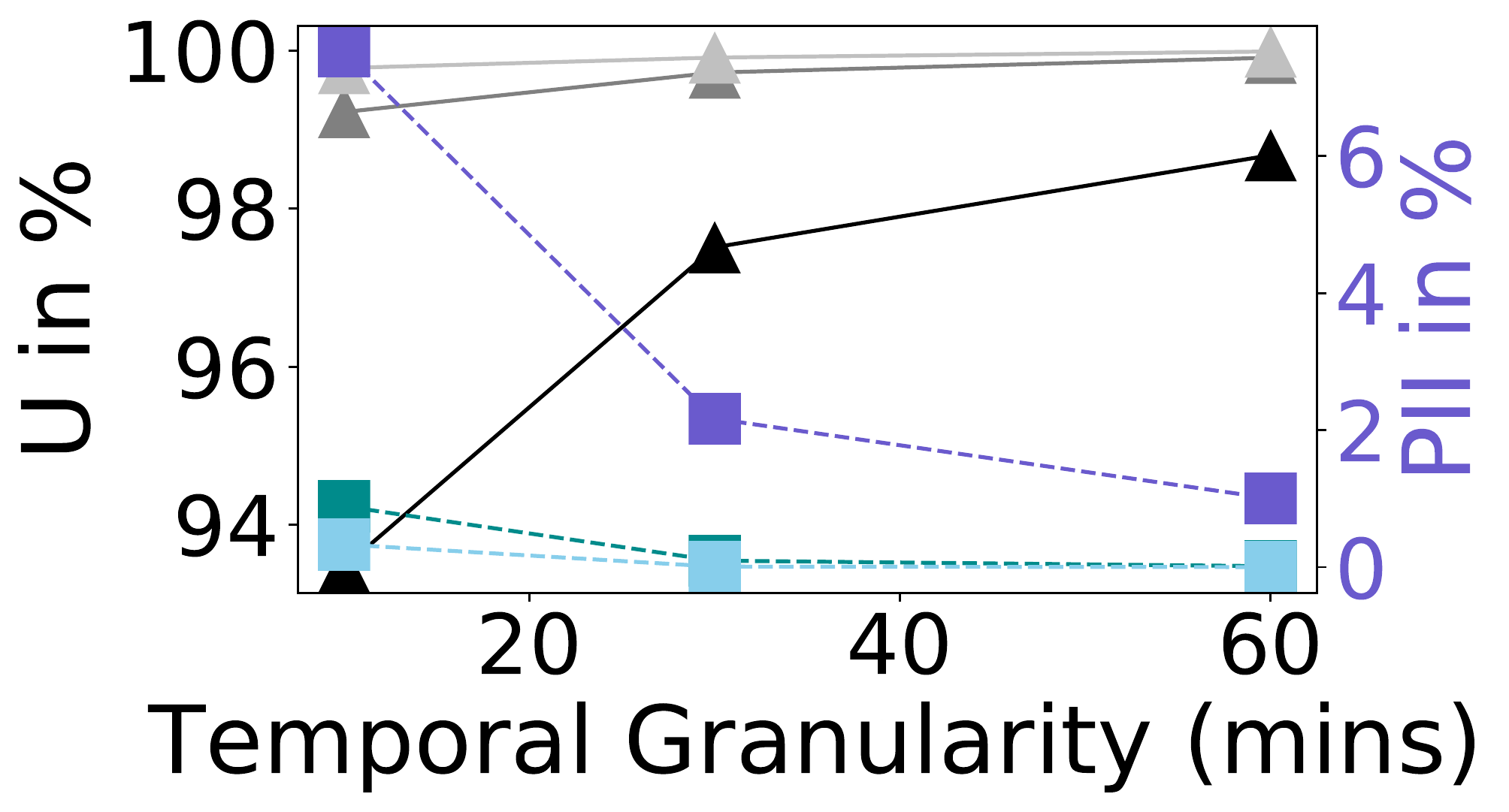}
         \caption{MDC}
         \label{fig:tem_mdc}
     \end{subfigure}
     \hfill
     \begin{subfigure}[b]{0.24\textwidth}
         \centering
         \includegraphics[width=\textwidth]{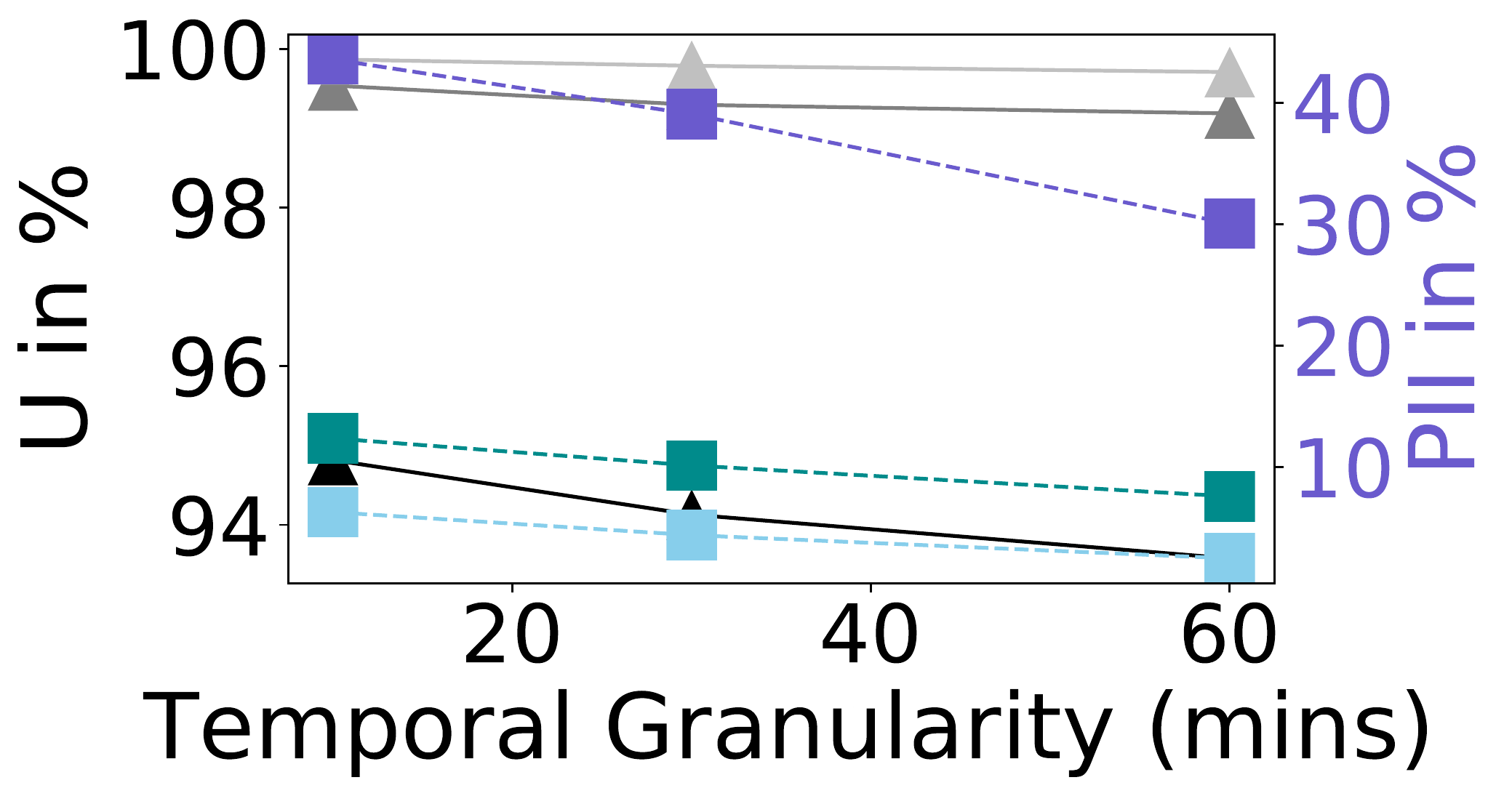}
         \caption{Priva\'Mov}
         \label{fig:tem_privamov}
    \end{subfigure}
    \hfill
    \begin{subfigure}[b]{0.24\textwidth}
         \centering
         \includegraphics[width=\textwidth]{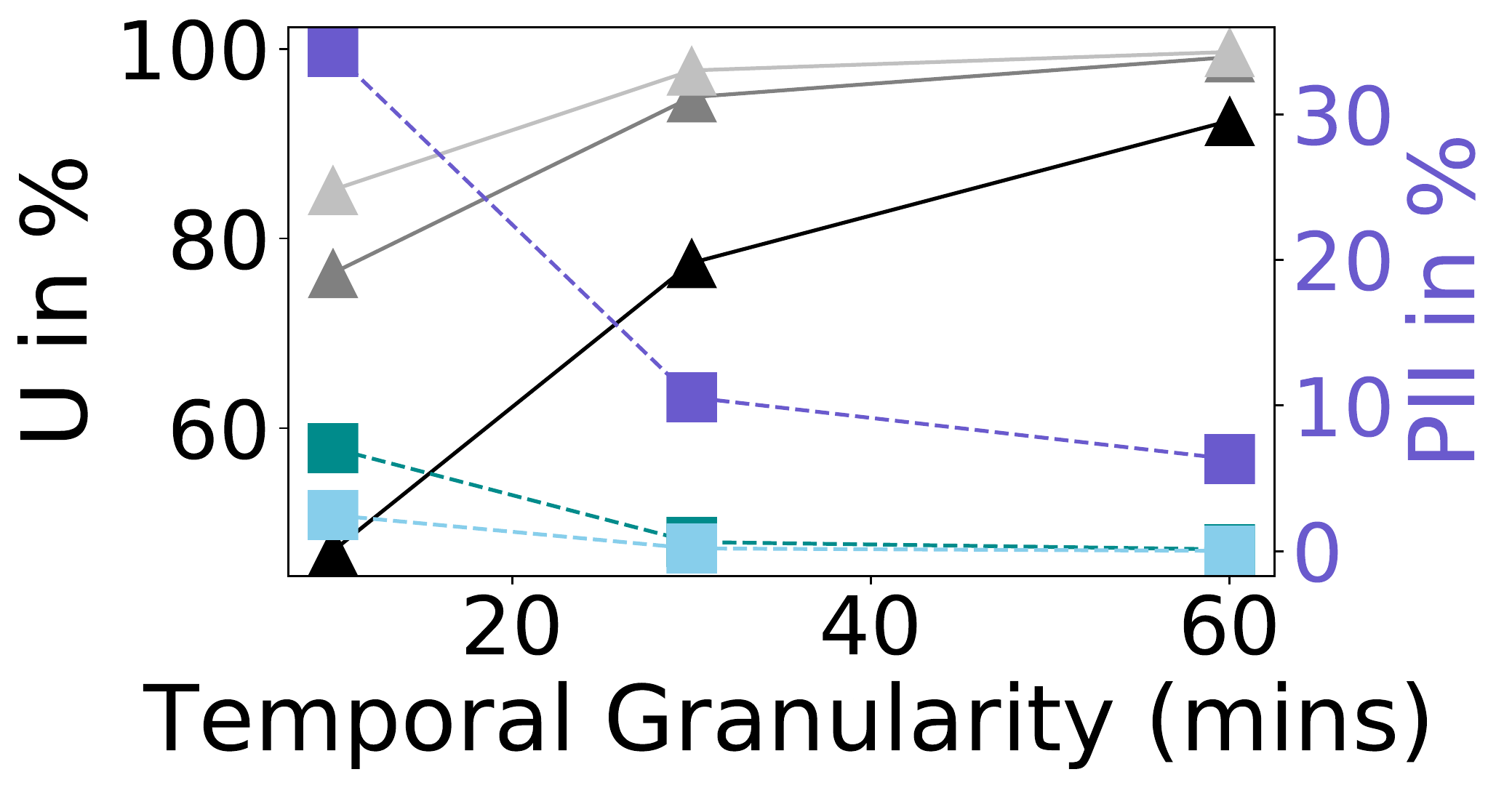}
         \caption{Geolife}
         \label{fig:tem_geolife}
     \end{subfigure}
     \hfill
     \begin{subfigure}[b]{0.24\textwidth}
         \centering
         \includegraphics[width=\textwidth]{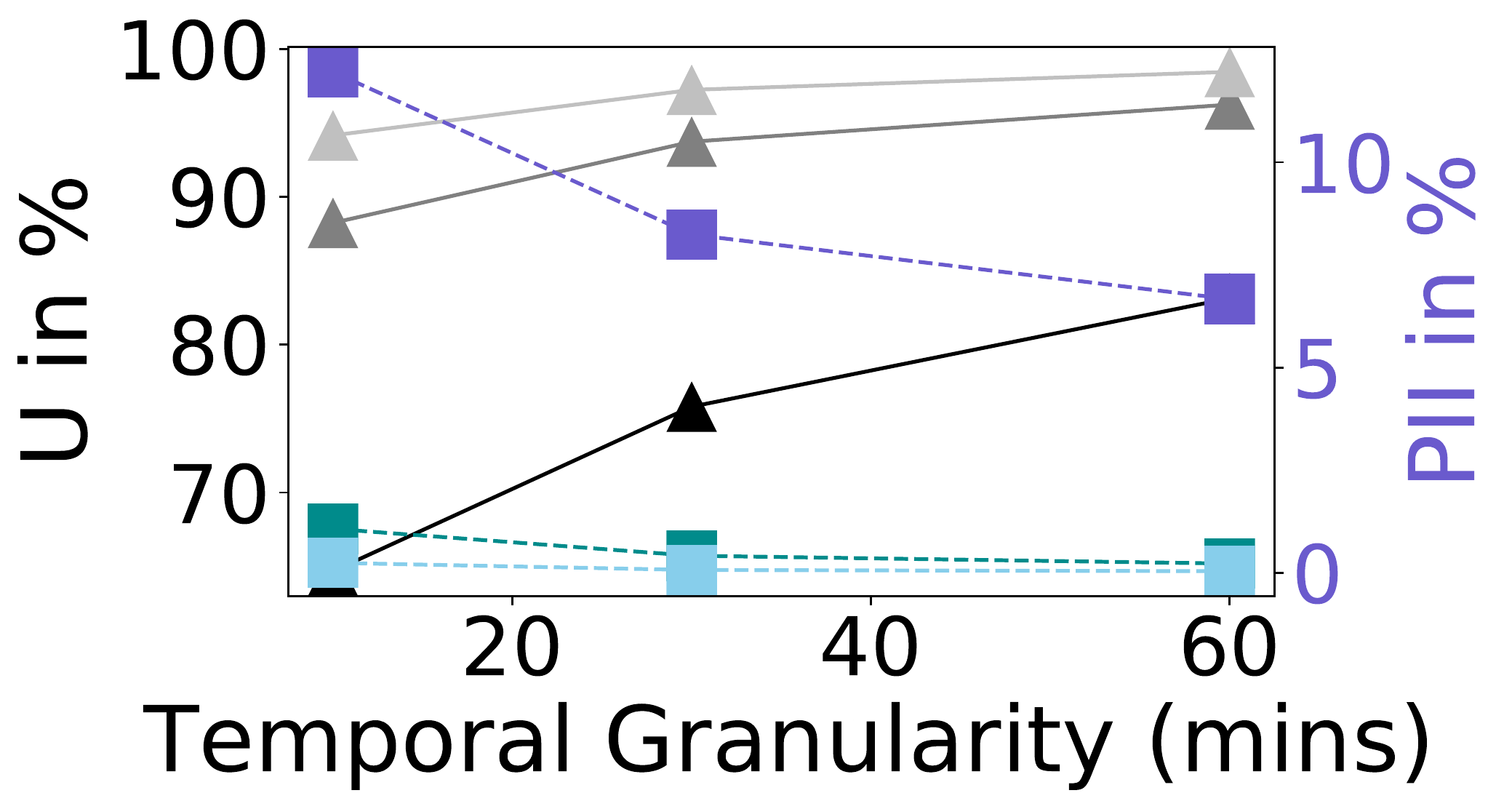}
         \caption{Foursquare}
         \label{fig:tem_foursquare}
    \end{subfigure}
    \hfill
    
    \caption{The effect of temporal granularity on the model performance of four mobility datasets.}
    \Description{}
    \label{fig:temporal}
\end{figure*}

Figure~\ref{fig:utility-privacy} presents \textit{Utility} and \textit{Privacy II} (\textit{i.e.}, user re-identification risk) trade-offs of the proposed system on the four datasets. 
The \textit{Zero} line (\textit{i.e.}, y = 0\%) in each sub-figure is leveraged to indicate the original utility rate (\textit{U}) and privacy rate (\textit{PII}) of the raw data. 
The blue line with square marker is the privacy gain rate with top-1 accuracy and the blue line without marker is the top-5 accuracy. 
The orange lines with and without triangle marker present the utility decline rate with top-1 and top-5 accuracies, respectively.
Hence, the orange area represents the utility decline while the light-green area represents the privacy gain when compared with original results. 
The dark-green area represents the trade-offs between utility and privacy budgets. 
The x-axis shows five different settings of the model, and the y-axis shows the trade-offs (\textit{i.e.}, \textit{trade-offs} = \textit{privacy gain} + \textit{utility decline}).

In summary, these trade-offs are all positive in different model settings on four different datasets. 
The performance on the Geolife data is the best, while less than 20\% utility decline but more than 50\% privacy gains. 
The performance on MDC and FourSquare also show the promising utility-privacy trade-offs, especially for setting \textit{V} on the FourSquare dataset, both the utility and privacy increase.
The uniqueness of human mobility trajectories is high, and these trajectories are likely to be re-identified even with a few location data points~\cite{de2013unique}.
Our results emphasize that the concern of user re-identification risk could be alleviated effectively with our proposed model.

\subsection{Discussion of Temporal Granularity}

The timestamp is one of the basic components of the trajectory sequence, and different choices on the temporal granularity affect the final performance of any dataset. 
Figure~\ref{fig:temporal} shows the impact of the varying temporal granularity on the proposed architecture.
We specifically present the top-1, top-5, and top-10 accuracies for both utility and privacy dimensions, respectively. 
For instance, when temporal granularity is 10-min, it indicates a location record \textit{r} is taken by every 10 minutes from the raw data. 
When using more coarser temporal granularity, the quantity of interested location points decreases, so as the difficulty of mobility prediction. 
However, the uniqueness of trajectory decreases due to ignoring many of the unique locations from each user, resulting in the lower privacy.
To summarize on the Figure~\ref{fig:temporal},
the impact of temporal granularity on the Priva'Mov is minimal. 
In terms of utility (mobility prediction), Priva'Mov is the only dataset for which accuracy decreases with increasing temporal granularity. This subtle decline emphasizes the trajectory features only has a small change when varying granularity, in line with the university students' mobility.

\section{Related work}
\label{RelW}

To contextualize our work, we briefly review the current state of the art in machine learning-based privacy preservation techniques for mobility data.

Current location privacy protection studies focus on two research streams.
One is the differential privacy approach to grouping and mixing the trajectories from different users so that the identification of individual trajectory data is converted into a k-anonymity problem~\cite{aktay2020google,xiao2015protecting,andres2013geo}. 
The other stream focuses on synthetic data generation~\cite{rezaei2018protecting,huang_variational_2019,choi2021trajgail, ijcai2018-530}. 
Synthetic data generation methods have been extensively studied in recent years as a way of tackling privacy concerns of location based datasets. 
The majority of existing mobility synthesis schemes are mainly categorized into two approaches:
One is a more traditional, simulation-based approach, while the other is a more recent, neural network based generative modeling approach that utilizes recurrent autoencoders and generative adversarial networks to produce realistic trajectories~\cite{shin2020user}.
Simulation based approaches generate mobility traces by modeling overall user behavior as a stochastic process, such as a Markov chain model of transition probabilities between locations, and then drawing random walks,
potentially with additional stochastic noise added, as demonstrated in Xiao et al \cite{xiao_loclok_2017}.
These approaches require considerable feature engineering effort and struggle to capture longer-range temporal and spatial dependencies in the data~\cite{luca2021survey} and are thus limited in their ability to preserve
the utility of the original datasets.
In contrast, the generative neural network approach synthesizes user mobility traces by learning, via gradient descent back-propagation, then the optimal weights are utilized for decoding a high-dimensional latent vector representation into sequences that closely resemble the original data. 
Such traces can maintain important statistical properties of the original data while taking advantage of noise introduced in the reconstruction process, to improve
data subject anonymity.
Huang et al \cite{huang_variational_2019} demonstrates the use of a variational autoencoder network to reconstruct trajectory sequences,
while Ouyang et al \cite{ijcai2018-530} utilizes a convolutional GAN, but neither work directly makes a quantitative
assessment of the extent of privacy protection that their algorithms provide \cite{huang_variational_2019, ijcai2018-530}.
The LSTM-TrajGAN by Rao et al \cite{rao2020lstm} is a state of the art example of the generative trajectory modeling approach, which quantifies its privacy protection by demonstrating a significant
decline in the performance of a second user ID classifier model on the synthetic outputs compared to the original input trajectories.
For these reasons, we used it as both a baseline for comparison and for the design of our proposed architecture.

Our proposed model takes the neural generative modeling approach, but differs from existing methods in that we utilize a combined, multi-task neural network to simultaneously reconstruct trajectories,
predict next locations, and reidentify users, from the same learned latent vector representation. 
We seek an optimal trade-off between the three tasks' individual losses by optimizing a Lagrangian loss function with per-task weights, improving the controllability of the relative utility and privacy of the outputs.

\section{Conclusion}
\label{Conl}

In this paper, we presented a privacy-preserving architecture based on the adversarial networks. Our model takes into account three different optimization objectives and searches for the optimum trade-off for utility and privacy of a given dataset. We reported an extensive analysis of our model performances and the impact of its hyper-parameters using four real-world mobility datasets.  
The Lagrange multipliers $\lambda_1,\ \lambda_2,$ and $\lambda_3$ bring more flexibility to our framework that enable it to satisfy different scenarios' requirements according to the relative importance of utility requirements and privacy budgets. 
We evaluated our framework on four datasets and benchmarked our results against an LSTM-GAN approach. The comparisons indicate the superiority of the proposed framework and the efficiency of the proposed privacy-preserving feature extractor $Enc_L$.
Expanding this work, we will consider other utility functions for our model such as community detection based on unsupervised clustering methods or deep embedded clustering methods. 
In future work, we will leverage automated search techniques, such as deep deterministic policy gradient algorithm, for efficiency in searching for the optimal Lagrange multipliers.


\bibliographystyle{ACM-Reference-Format}
\bibliography{ref}

\onecolumn
\newpage
\appendix
\textbf{APPENDIX}

In general, the encoder $Enc_L$ should satisfy high predictability (\textbf{\textit{min}}~$\mathcal{L_U}$) and low user re-identification accuracy (\textbf{\textit{max}}~$\mathcal{L_P}$) of the mobility data when maximizing the reconstruction loss (\textbf{\textit{max}}~$\mathcal{L_R}$) in the reversed engineering.
The overall training is to achieve privacy-utility trade-off by adversarial learning on $\mathcal{L_R}$, $\mathcal{L_U}$, and $\mathcal{L_P}$ concurrently.
The gradient of the loss (\textit{i.e.}, $\theta_R$, $\theta_U$, $\theta_P$) back-propagate through the LSTM network to guide the training of the encoder $Enc_L$.
The encoder is updated with the \textit{sum loss} function $\mathcal{L}_{sum}$ until convergence.
Algorithm 1 summarizes our training setting of the LSTM-PAE model.

\begin{algorithm*}[]

\SetKwInOut{Input}{Input}\SetKwInOut{Output}{Output}
\SetAlgoLined
	\Input{Mobility data \textbf{X}, real mobility prediction labels \textbf{Y}, real identification recognition labels \textbf{Z}, Lagrangian multipliers: $\lambda_1$, $\lambda_2$, $\lambda_3$} 
	\Output{Adversarial Encoder $Enc_L(X)$}
	Initialize model parameters\\
	 \For{n epochs}{ 
	 	\For{$k = 1$, $\cdots$$, K_t$}{
	 	 1. Sample a min-batch of mobility trajectories x, prediction labels y, identification labels z\\
	 	 2. Update $\theta_R$ with Adam optimizer on mini-batch loss $L_R(\theta_R)_{(x, \hat{x})}$\\
	 	 3. Update $\theta_U$ with Adam optimizer on mini-batch loss $L_U(\theta_U)_{(y, \hat{y})}$\\
	 	 4. Update $\theta_P$ with Adam optimizer on mini-batch loss $L_P(\theta_P)_{(z, \hat{z})}$\\
  	    }
  	    Update with the gradient descent on $L_{sum}(\theta_R, \theta_U, \theta_P, \lambda_1, \lambda_2, \lambda_3)$
 	 } 
    \caption{Training of the proposed adversarial network for representation learning}
    \label{algo_disjdecomp} 
\end{algorithm*}

\end{document}